\pdfoutput=1
\documentclass[11pt,table]{article}
\usepackage[utf8]{inputenc}         %
\usepackage[T1]{fontenc}            %
\usepackage{libertine}              %
\usepackage[scaled=0.85]{beramono}  %
\usepackage[in]{fullpage}           %
\usepackage[sc]{titlesec}           %
\usepackage{xurl}                   %
\usepackage{amsmath}                %
\usepackage{booktabs}               %
\usepackage[dvipsnames]{xcolor}     %
\usepackage{graphicx}               %
\graphicspath{{figures/}}           %
\usepackage{subcaption}             %
\usepackage{microtype}              %
\usepackage[square]{natbib}         %
\usepackage{siunitx}                %
\usepackage{listings}               %
\usepackage{algorithm}              %
\usepackage{algorithmic}            %

\usepackage[pdfusetitle]{hyperref}  %
\usepackage[nameinlink]{cleveref}   %
\newcommand\myshade{90}
\definecolor{mylinkcolor}{HTML}{8F510B} %
\definecolor{mycitecolor}{HTML}{3B4791} %
\definecolor{myurlcolor}{HTML}{8F7F0B}  %
\hypersetup{
    linkcolor  = mylinkcolor!\myshade!black,
    citecolor  = mycitecolor!\myshade!black,
    urlcolor   = myurlcolor!\myshade!black,
    colorlinks = true,
}

\newcommand{\inum}[1]{(\emph{#1})}
\newcommand{\tinysection}[1]{

    $ $

    \noindent
    \textbf{#1.} 
}
\newcommand{\SET}[1]{\mathcal{#1}} %
\newcommand{\REALSET}[1]{\mathcal{#1}} %
\newcommand{\VEC}[1]{\boldsymbol{#1}} %
\newcommand{\MAT}[1]{\boldsymbol{#1}}
\newcommand{\VECVAR}[1]{\text{\textbf{\emph{#1}}}} %

\begin{document}

\title{\bf\sc ClipUp: A Simple and Powerful Optimizer for Distribution-based Policy Evolution}
\author{Nihat Engin Toklu \and Pawe\l{} Liskowski \and Rupesh Kumar Srivastava}
\date{
    \{ engin | pawel | rupesh \}@nnaisense.com\\
    \vspace{0.3cm}
    NNAISENSE Technical Report\footnote{Extends work presented at the International Conference on Parallel Problem Solving from Nature 2020. The final authenticated publication is available online at \url{https://doi.org/10.1007/978-3-030-58115-2_36}.}\\
    Code available at: \url{https://github.com/nnaisense/pgpelib}
}

\maketitle

\begin{abstract}
Distribution-based search algorithms are an effective approach for evolutionary reinforcement learning of neural network controllers.
In these algorithms, gradients of the total reward with respect to the policy parameters are estimated using a population of solutions drawn from a search distribution, and then used for policy optimization with stochastic gradient ascent.
A common choice in the community is to use the Adam optimization algorithm for obtaining an adaptive behavior during gradient ascent, due to its success in a variety of supervised learning settings.
As an alternative to Adam, we propose to enhance classical momentum-based gradient ascent with two simple techniques: gradient normalization and update clipping.
We argue that the resulting optimizer called ClipUp (short for \emph{clipped updates}) is a better choice for distribution-based policy evolution because its working principles are simple and easy to understand and its hyperparameters can be tuned more intuitively in practice.
Moreover, it removes the need to re-tune hyperparameters if the reward scale changes.
Experiments show that ClipUp is competitive with Adam despite its simplicity and is effective on challenging continuous control benchmarks, including the Humanoid control task based on the Bullet physics simulator.
\end{abstract}

\section{Introduction}
\label{sec:introduction}

We propose a simple and competitive optimizer (an adaptive gradient following mechanism) for use within distribution-based evolutionary search algorithms for training reinforcement learning (RL) agents.
Distribution-based search \citep{hansen1997,salomon1998,hansen2001,salomon2005,sehnke2010,wierstra2014,salimans2017,mania2018} is a simple but powerful category of evolutionary algorithms. The common principles of distribution-based search algorithms can be summarized as follows:

$ $
\begin{tabular}{l p{14cm}}
    \textbf{0:} & Initialize the current solution.\\
    \textbf{1:} & Sample neighbor solutions from a search distribution centered at the current solution.\\
    \textbf{2:} & Evaluate each neighbor solution, estimate a gradient which is in the direction of the weighted average (in terms of solution fitnesses) of the neighboring solutions.\\
    \textbf{3:} & Update the current solution using the gradient.\\
    \textbf{4:} & Go to 1.\\
\end{tabular}

$ $

In step 3, it is possible to use any stochastic gradient ascent algorithm to potentially speed up policy optimization through the use of adaptive update algorithms  such as momentum \citep{polyak1964} or Adam \citep{kingma2015}.
These adaptive optimizers are commonly used in the supervised learning community since they take into account not just the current gradient, but often its previous values as well to compute a more informed update.
Following its success in supervised learning, the Adam optimizer in particular has been commonly used in recent work on neuroevolutionary RL \citep{salimans2017,ha2019,freeman2019}.

Using adaptive optimizers instead of plain gradient ascent can potentially speed up training, but confronts the practitioner with new challenges.
A basic question to consider is whether optimizers designed to work in the supervised learning setting (where gradients are computed by differentiation) sufficiently address the specific issues that arise when training RL agents with distribution-based search.
Secondly, these optimizers often introduce several additional hyperparameters, all of which must be tuned in order to capitalize on their abilities even for supervised learning, as demonstrated empirically by \citet{choi2019empirical}.
But many hyperparameters are non-intuitive to tune in our setting, and most practitioners instead tune one or two primary hyperparameters using a few trials while keeping the rest at their default values.
This potentially leaves performance gains on the table.

Focusing on the practitioner's experience, we contribute a solution to these issues with a general-purpose adaptive optimizer that is especially suitable for embedding into the framework of distribution-based search for RL.
\emph{ClipUp}, short for ``clipped updates'', combines a few simple techniques: stochastic gradient ascent with momentum, gradient normalization, and update clipping.
We argue that it is a valuable tool for RL practitioners because its hyperparameters are very easy to understand, providing valuable intuitions for tuning them for a given problem.
In a series of continuous control experiments, we compare it to Adam and show that
\inum{i} ClipUp is insensitive to reward function rescaling while Adam needs to be re-tuned for each scale; and
\inum{ii} ClipUp performs on par with Adam on \texttt{Walker2d-v2} and \texttt{Humanoid-v2} robot control benchmarks based on the Mujoco simulator.
Finally, we demonstrate that ClipUp can also solve the PyBullet \citep{coumans2019} humanoid control task, a challenging RL environment reported to be much harder \citep{coumans2018note} than its MuJoCo counterpart.

\section{Background}
\label{sec:background}

\subsection{Policy gradients with parameter-based exploration}
\label{sec:pgpe}

All experiments in this study use \emph{policy gradients with parameter-based exploration} (PGPE; \citealp{sehnke2010}) as a representative distribution-based search algorithm.
A variant of PGPE was also used by \citet{ha2017,ha2019}, demonstrating that it can be successful on recent RL benchmarks.
Although PGPE draws inspiration from an RL-focused study \citep{williams1992}, it is a general-purpose derivative-free optimization algorithm, and can be considered to be a variant of evolutionary gradient search algorithms \citep{salomon1998,salomon2005}.

The PGPE algorithm is described in \Cref{alg:pgpe}.
Each iteration of PGPE works as follows.
First (line 2), a new population is built by sampling neighbor solutions around the current solution $\VEC{x}_k$.
These neighbors are sampled from a Gaussian distribution whose shape is expressed by the standard deviation vector $\VEC{\sigma}_k$.
Following \citet{salomon2005}, solutions are sampled symmetrically: when a new neighbor solution $\VEC{x}_k+\VEC{\delta}$ is added to the population, its \emph{mirror} counterpart $\VEC{x}_k-\VEC{\delta}$ is added as well.
We denote our population of directions sampled at iteration $k$ as $\SET{D}_k$.
The next step (line 3) of the algorithm is to find the gradient for updating the current solution, which is computed by the weighted average of all the fitness gains along the directions.
The algorithm then (line 4) computes the gradient for updating the standard deviation vector.
Finally these gradients are used for computing the new current solution and the new standard deviation (line 5).

\begin{algorithm}[htb]
    \caption{The PGPE algorithm \citep{sehnke2010}}
    \label{alg:pgpe}

    \begin{tabular}{l l}
        \textbf{Hyperparameters:} & Population size $\lambda$ \;\;(expected as an even number)\\
                                  & Initial solution $\VEC{x}_0$ \;\;(in our study, set as near zero) \\
                                  & Initial standard deviation vector $\VEC{\sigma}_0$ \\
                                  & Standard deviation learning rate $\Omega$ \\
                                  & $\text{AdaptiveOptimizer} \in \{\text{Adam}, \text{ClipUp}\}$ \\
    \end{tabular}

    \begin{algorithmic}[1]
        \FOR{iteration $k=0,1,2,...$}

        \STATE Build a population of directions
        $$
        \begin{array}{l l l l l}
            \SET{D}_k \gets \big\{ & (\VEC{d}^+_i, \VEC{d}^-_i) & \big|  & \VEC{d}^+_i = \VEC{x}_k + \VEC{\delta}_i, \\
                             &                &        & \VEC{d}^-_i = \VEC{x}_k - \VEC{\delta}_i, \\
                             &                &        & \VEC{\delta}_i \sim \big(\REALSET{N}(0, \MAT{I}) \cdot \VEC{\sigma}_k \big), \\
                             &                &        & i \in \{1, 2, ..., \lambda / 2 \} & \big\}
        \end{array}
        $$

        \STATE Estimate the gradient for updating the current solution $\VEC{x}$
        $$
        \nabla_{\VEC{x}_k} \gets \frac{1}{|\SET{D}_k|} \cdot \sum_{(\VEC{d}^+,\VEC{d}^-) \in \SET{D}_k} \bigg[ \frac{f(\VEC{d}^+) - f(\VEC{d}^-)}{2} \cdot (\VEC{d}^+-\VEC{x}_k) \bigg]
        $$

        \STATE Estimate the gradient for updating the standard deviation vector $\VEC{\sigma}$
        $$
        \nabla_{\VEC{\sigma}_k} \gets
        \frac{1}{|\SET{D}_k|}
        \cdot
        \sum_{(\VEC{d}^+,\,\VEC{d}^-) \in \SET{D}_k}
        \Bigg[
        \bigg(
        \frac{f(\VEC{d}^+) + f(\VEC{d}^-)}{2} - \widetilde{f}
        \bigg)
        \cdot
        \bigg(
        \frac{(\VEC{d}^+-\VEC{x}_k)^2 - (\VEC{\sigma}_k)^2}{\VEC{\sigma}_k}
        \bigg)
        \Bigg]
        $$
        \quad\quad\quad\quad\quad {\small where $\widetilde{f}$ is the average fitness of all the solutions in $\SET{D}_k$, serving as a baseline value}

        \STATE Perform the updates
        $$
        \begin{array}{r c l}
            \VEC{x}_{k+1} &\gets& \VEC{x}_k + \text{AdaptiveOptimizer}(\nabla_{\VEC{x}_k}) \\
            \VEC{\sigma}_{k+1} &\gets& \VEC{\sigma}_k + \Omega \cdot \nabla_{\VEC{\sigma}_k}
        \end{array}
        $$

        \ENDFOR
    \end{algorithmic}

    {\footnotesize
    \quad\quad\quad\quad
    Division between two vectors, and squaring of a vector are elementwise operations
    }

\end{algorithm}

\citet{ha2017,ha2019} previously enhanced PGPE in three ways:
\inum{i} solutions were fitness-ranked (from worst to best, the ranks range linearly from -0.5 to 0.5) and their ranks were used for gradient computations instead of their raw fitnesses;
\inum{ii} the Adam optimizer was used for following the gradients in an adaptive manner;
\inum{iii} to make sure that the standard deviation updates remain stable, 
the updates for the standard deviation were clipped in each dimension to 20\% of their original values.
\inum{i} and \inum{ii} were also previously shown to be successful in the evolution strategy variant studied by \citet{salimans2017}.
We adopt these enhancements in this study and incorporate two further RL-specific enhancements listed below.

$ $

\noindent
\textbf{Adaptive population size} \citep{salimans2017}.
When considering locomotion problems where the agent bodies are unstable,
wrong actions cause the agents to fall, breaking constraints and ending the trajectories abruptly.
In the beginning, most of the agents fall immediately.
Therefore, to find reliable gradients at the beginning of the search, very large populations are required so that they can explore various behaviors.
However, such huge populations might be unnecessary once the search finds a reliable path to follow.
Therefore, in addition to the population size $\lambda$, we use a hyperparameter $T$, which is the total number of environment time steps (i.e. number of interactions done with the simulator) that must be completed within an iteration.
If, after evaluating all the solutions within the population, the total number of timesteps is below $T$, the trajectories are considered to be too short (most agents fell down) and the current population size is increased (by $\lambda$ more solutions in our implementation) until the total number of environment timesteps reaches $T$, or the extended population size reaches an upper bound $\lambda^{\text{max}}$.
This mechanism results in an automatic decay of the population size during the evolution process.

$ $

\noindent
\textbf{Observation normalization} \citep{salimans2017,mania2018}.
We normalize observations using the running statistics over all the observations received by all the agents until the current iteration.

In the remainder of this paper, we use the notation PGPE+ClipUp to refer to PGPE combined with ClipUp as the adaptive gradient following algorithm.
Similarly, we use the notation PGPE+Adam for when Adam is used instead of ClipUp.

\subsection{Heavy ball momentum}
The heavy ball method \citep{polyak1964} is a very early momentum-based optimizer for speeding up convergence of gradient descent. 
Considering the current solution as a ball moving in the solution space, each gradient contributes to the velocity of this ball. 
This means that the directions consistently pointed to by the recent gradients are followed more confidently (because the velocity accumulates in those directions), and similarly, directions rarely pointed to are followed more cautiously (or instead, they just contribute negatively to the current velocity to some extent).

When using distribution-based evolutionary search algorithms, the gradients can be very noisy because
\inum{i} they are estimated stochastically using a sampled population;
and \inum{ii} the objective function is a simulator which itself might be stochastic (e.g. because the simulator is a physics engine relying on stochastic heuristics, or it deliberately injects uncertainty to encourage more robust policies).
The concept of momentum can be useful when dealing with noisy gradients, because, the velocity will accumulate towards the historically consistent components of the noisy gradients, and misleading inconsistent components of the gradients will cancel out.

Note that the Adam optimizer inherits the concept of momentum as well.
Evolution strategy with covariance matrix adaptation  \citep[CMA-ES;][]{hansen1997,hansen2001} also implements a variant of the momentum mechanism called ``evolution path''.

\subsection{Gradient normalization}
Used by \citet{salomon1998} in the context of evolutionary search, gradient normalization has the useful effect of decoupling the direction of a gradient and its magnitude. The magnitude of the gradient can then be re-adjusted or overwritten by another mechanism, or simply by a constant.

When there is no gradient normalization, the magnitude of a gradient would be computed as a result of the weighted average performed over the fitness values of the population. The most important problem with unnormalized gradients is that one has to tune the step size according to the scales of the fitness values, which vary from problem to problem, or even from region to region within the solution space of the same RL problem. To counter the varying fitness scale issue, one can employ fitness ranking as done in prior work \citep[e.g.][]{hansen2001,wierstra2014,salimans2017,ha2019}. However, even then, the step size must be tuned according to the scale imposed by the chosen fitness ranking method.

On the other hand, let us now consider the simple mechanism of normalizing the gradient as $\alpha \cdot (\VEC{g} \, / \, ||\VEC{g}||)$, where $\alpha$ is the step size, $\VEC{g}$ is the unnormalized gradient, and $||\VEC{g}||$ is the $L_{2}$ norm of $\VEC{g}$. With this mechanism, the step size $\alpha$ becomes a hyperparameter for tuning the Euclidean distance expressed by the normalized gradient, independent of the scale of the fitness values or ranks.
In addition to the advantage of being scale independent, we argue that with this mechanism, it is easy to come up with sensible step size values for updating a policy.

\subsection{Gradient clipping}
In the supervised learning community, gradient clipping by value \citep{mikolov2012a,graves2013a} or by norm \citep{pascanu2013} (see also \citet{zhang2020} for a recent analysis) has become common practice for avoiding instabilities due to exploding gradients \citep{hochreiter1991}.
The technique used in this paper is related but slightly different. 
We clip the updated velocity of the heavy ball (just before updating the current solution), which is why we call it \emph{update clipping}.
It works as follows: if the norm of the velocity is larger than a maximum speed threshold, then the velocity is scaled such that its magnitude is reduced to the threshold, but its direction remains unchanged.
The intuition behind clipping the velocity of the heavy ball method is to prevent it from gaining very large velocities that can overshoot the (local) optimum point.

\begin{algorithm}[t]
    \caption{The ClipUp optimizer}
    \label{alg:clipup}

    \begin{tabular}{r l}
        \textbf{Initialization:}  & Velocity $\VEC{v}_0 = \VEC{0}$ \\
        \textbf{Hyperparameters:} & Step size $\alpha$ \\
                                  & Maximum speed $v^{\text{max}}$ \\
                                  & Momentum $m$ \\
        \textbf{Input:}           & Estimated gradient $\nabla f(\VEC{x}_k)$
    \end{tabular}

    $ $

    \begin{algorithmic}[1]
        \STATE $\VEC{v'}_{k+1} \gets m \cdot \VEC{v}_k + \alpha \cdot \big( \nabla f(\VEC{x}_k) \,/\, ||\nabla f(\VEC{x}_k)|| \big)$
        \IF{$||\VEC{v'}_{k+1}|| > v^{\text{max}}$}
            \STATE $\VEC{v}_{k+1} \gets v^{\text{max}} \cdot \big( \VEC{v'}_{k+1} \,/\, ||\VEC{v'}_{k+1}|| \big)$
        \ELSE
            \STATE $\VEC{v}_{k+1} \gets \VEC{v'}_{k+1}$
        \ENDIF
        \STATE \textbf{return } $\VEC{v}_{k+1}$
    \end{algorithmic}

\end{algorithm}

\section{The ClipUp Optimizer}
\label{sec:clipup}

We now discuss the ClipUp optimizer, which is a combination of the heavy ball momentum, gradient normalization, and update clipping techniques discussed in \Cref{sec:background}.

Let us consider an optimization problem with the goal of maximizing $f(\VEC{x})$, where $\VEC{x}$ is a solution vector.
We denote the gradient of $f(\VEC{x})$ as $\nabla f(\VEC{x})$.
In the context of evolutionary RL, it is usually the case that $f(\VEC{x})$ is not differentiable, therefore, it is estimated by using the fitness-weighted (or rank-weighted) average of the population of neighboring solutions.

When using gradient ascent without any adaptive optimizer, at iteration $k$ with step size $\alpha$, the following simple update rule would be followed:
\begin{equation*}
    \VEC{x}_{k+1} \gets \VEC{x}_k + \alpha \cdot \nabla f(\VEC{x}).
\end{equation*}
With ClipUp, the update rule becomes
\begin{equation*}
    \VEC{x}_{k+1} \gets \VEC{x}_k + \text{ClipUp}\big(\nabla f(\VEC{x})\big),
\end{equation*}
where ClipUp is defined in \Cref{alg:clipup}.
First (line 1), the algorithm normalizes the gradient and scales it by the step size $\alpha$, fixing the gradient's magnitude to $\alpha$.
Then it computes a new velocity by adding the $\alpha$-sized gradient to the decayed velocity of the previous iteration (where decaying means that the previous velocity is scaled by the momentum factor $m$, usually fixed to 0.9).
The next step of the algorithm (lines 2 to 6) is to clip this newly computed velocity if its magnitude exceeds the threshold imposed by the hyperparameter $v^{\text{max}}$.
When clipped, the velocity's magnitude is reduced to $v^{\text{max}}$; its direction remains unchanged.
Finally, the procedure ends by returning the clipped velocity (line 7).
Note that ClipUp clips the velocity (or solution \emph{update}) instead of clipping the gradient.
In the context of training deterministic control policies, this is a simple way of preventing the policy from changing too much during a single update.

With the normalization and the clipping operations employed by ClipUp, the two hyperparameters $\alpha$ and $v^{\text{max}}$ have intuitive interpretations and can be tuned on the same scale as the perturbation one would like to apply on the current solution.
The step size $\alpha$ is now the fixed norm of the gradient vector that updates the velocity, and the maximum speed $v^{\text{max}}$ expresses the maximum norm of the change possible to the current solution vector.
They are completely independent of the fitness scale of the problem or the fitness-based ranking employed on the population of solutions, which otherwise affects the approximated gradient.

Like Adam, ClipUp does not make any assumptions about the search algorithm employing it. Therefore, although we use PGPE as our search algorithm, in theory it is possible to use ClipUp with other similar evolutionary algorithms as well, such as the evolution strategy variant used by \citet{salimans2017}.

\section{Hyperparameter Tuning for PGPE+ClipUp}
\label{sec:tuning}

Having the step size $\alpha$ and the maximum speed $v^{\text{max}}$ on the same scale as the solution perturbation allows us to come up with simple-yet-effective heuristic scheme for tuning hyperparameters when solving a variety of control problems.
The procedure is as follows: we designate $v^{\text{max}}$ as the main hyperparameter to be tuned.
Its value simply bounds the maximum possible change in the norm of the solution vector (which happens when the solution vector is fully aligned with the velocity) and in some cases can be intuitively tuned by a practitioner.
The step size $\alpha$ is set simply to $v^{\text{max}} / 2$, and $m$ is always set to 0.9.
This completely specifies the ClipUp hyperparameters.

\begin{figure}
    \centering
    \includegraphics[width=0.6\textwidth]{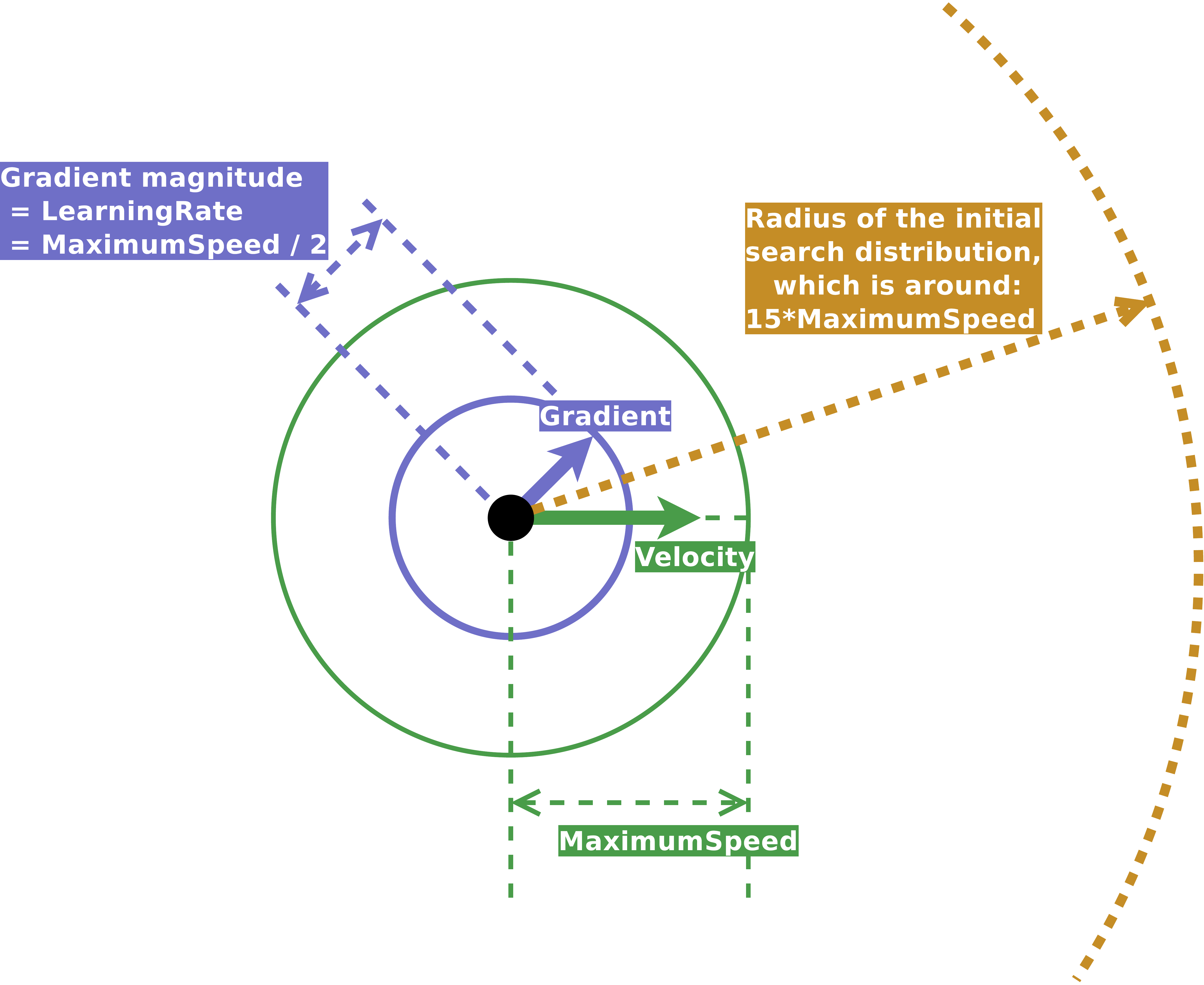}
    \caption{Visualization of the default hyperparameters for ClipUp. See \Cref{sec:tuning} for a recommended hyperparameter tuning procedure that we have found to be economical and reliable in practice.}
    \label{fig:defaultsettings}
\end{figure}

Finally, we remove the need to search for the important PGPE hyperparameter $\VEC{\sigma}_0$ by extending our intuitions from ClipUp.
Recall that $\VEC{\sigma}_0$ specifies the initial standard deviation vector for the Gaussian noise used to generate neighboring solutions.
Instead of directly setting $\VEC{\sigma}_0$, we define the \emph{radius} of the search distribution $r = ||\VEC{\sigma}_0||$.
This again brings us to the same distance scale on which $\alpha$ and $v^{\text{max}}$ are defined, and we set $\VEC{\sigma}_0$ such that $r = q \cdot v^{\text{max}}$, where values of $q$ between 10 and 20 generally work well, and a good default value is 15.
The relationships between the default hyperparameter settings above are visualized in \Cref{fig:defaultsettings}.

The above scheme for setting default hyperparameters has two important advantages for practical use: 
\inum{i} In many cases (and all problems considered in this paper), tuning a single hyperparameter is sufficient. 
\inum{ii} If one wishes to tune other hyperparameters, they have intuitive related interpretations that can help guide the search.
Note the contrast to Adam, where the hyperparameters $\epsilon, \beta_1, \beta_2, \alpha$ all have different interpretations and can not be easily adjusted with respect to each other.

Although we observed that the heuristic hyperparameter tuning rules mentioned here are effective on their own (as will be seen in \Cref{sec:experiments}), one can start with these rules and then fine-tune the step size and the radius to further increase the performance of PGPE+ClipUp on the task at hand.
In \Cref{sec:tunestepsize}, we demonstrate a case of such fine-tuning by increasing the step size $\alpha$ towards $v^{\text{max}}$.

\section{Experiments}
\label{sec:experiments}

In this section, we present the results obtained with PGPE+ClipUp with comparisons against PGPE+Adam.
For ClipUp, the tuning heuristics proposed in  \Cref{sec:tuning} were applied:
$\alpha=v^{\text{max}}/2$, $r=15 \cdot v^{\text{max}}$ (18 instead of 15 in the case of \texttt{Humanoid-v2}).
For Adam, we adopted the following hyperparameter values from the original paper \citep{kingma2015}: $\beta_1=0.9$, $\beta_2=0.999$, and $\epsilon=\num{1e-8}$ (the same default values were used by \citet{salimans2017}).
A single episode of interaction with the environment was used to compute $f$ during training.
For testing the current solution at any point, the average return over 16 episodes was recorded for reporting in tables and plots.
When a comparison is made between PGPE+ClipUp and PGPE+Adam, we use our best known search distribution radius for both, and tune the step size of Adam for each RL environment.

\subsection{Fitness scale (in)sensitivity}\label{subsec:scale}

We argued that the most important factor contributing to the intuitiveness of ClipUp is that its step size is configured directly in terms of the mutation magnitude. The normalization operator employed within ClipUp ensures that this step size configuration is not affected by the fitness scale.
To support this argument, we now compare the behaviors of PGPE+ClipUp and PGPE+Adam on the RL environment \texttt{LunarLanderContinuous-v2} using multiple fitness scales.
The environment has 8-dimensional observations and 2-dimensional actions.
We compare four setups, leading to four different fitness scales for the same task:
\inum{i} 0-centered fitness ranking;
\inum{ii} raw (original) reward values;
\inum{iii} raw reward values multiplied by 1000; and
\inum{iv} raw reward values divided by 1000.

\begin{table}
    \caption{Hyperparameters used for \texttt{LunarLanderContinuous-v2}}
    \label{tbl:hyperllc}
    \centering
    \begin{tabular}{c | c | c}
        \toprule
        PGPE & ClipUp & Adam \\
        \midrule
        $
        \begin{array}{rcl}
            \lambda &=& 200 \\
            \Omega &=& 0.1 \\
            r &=& v^{\text{max}} \cdot 15 \,=\, 4.5 \\
            \text{ObNorm}^{(a)} &=& \text{No} \\
        \end{array}
        $
        &
        $
        \begin{array}{rcl}
            v^{\text{max}} &=& \num{0.3} \\
            \alpha &=& v^{\text{max}} / 2 \,=\, \num{0.15} \\
            m &=& 0.9 \\
        \end{array}
        $
        &
        $
        \begin{array}{rcl}
            \alpha &\in& \{0.15, 0.175, 0.2\} \,\,^{(b)} \\
            \beta_1 &=& 0.9 \\ 
            \beta_2 &=& 0.999 \\
            \epsilon &=& \num{1e-8} \\
        \end{array}
        $ \\
        \bottomrule
    \end{tabular}
    
    $ $

    { \footnotesize
    (a) Observation normalization \quad
    (b) Tuned in the set $\{0.1, 0.125, 0.15, 0.175, 0.2\}$ (10 runs for each step size)
    }
    
\end{table}

For each reward scale, each optimizer algorithm, and each step size for Adam, we ran 10 experiment runs, each training a policy for 50 iterations.
We used a simple linear policy, which can be expressed as
$\VECVAR{act} = \VECVAR{obs} \cdot \MAT{W}$, where $\VECVAR{act}$ is the action vector generated by the policy, $\VECVAR{obs}$ is the observation vector received by the agent, and $\MAT{W}$ is a $8 \times 2$ sized weight matrix.
As $\MAT{W}$ represents the trainable parameters of the policy, we have 16 optimization variables.
The hyperparameters are reported in \Cref{tbl:hyperllc}.
The overall score of each group of 10 experimental runs was finally recorded as the mean over the results. These results are shown in \Cref{tbl:lunarlander}.

\begin{table}
    \caption{Comparison of PGPE+ClipUp and PGPE+Adam across four reward scales on LunarLanderContinuous-v2. Values outside the parentheses represent the final score, averaged across 10 runs.
    Values inside parentheses are the scores expressed as percentages of the same method's score when using fitness ranking.}
    \label{tbl:lunarlander}
    \centering
    
    \begin{tabular}{r c c c c}
        \toprule
                                       & ClipUp            & Adam ($\alpha=0.15$) & Adam ($\alpha=0.175$) & Adam ($\alpha=0.2$) \\
        \midrule
        Fitness ranking                & 269.95 (100.00\%) & 255.32 (100.00\%) & 241.94 (100.00\%) & 263.73 (100.00\%)\\ 
        Raw rewards                    & 270.15 (100.07\%) & 197.93 \;(77.52\%) & 235.72 \;(97.43\%) & 199.25 \;(75.55\%)\\ 
        Rewards $\times$ 1000     & 262.98 \;(97.42\%) & 139.61 \;(54.68\%) & 211.84 \;(87.56\%) & 245.67 \;(93.15\%)\\ 
           Rewards $/$ 1000     & 263.06 \;(97.44\%) & 200.34 \;(78.46\%) & 187.25 \;(77.39\%) & 111.76 \;(42.38\%)\\
        \bottomrule
    \end{tabular}
    
    {\footnotesize
    \begin{tabular}{p{11cm}}
    \end{tabular}
    }
\end{table}

It can be seen from the table that ClipUp is not affected at all by various reward scales. 
The small amount of deviation observed for ClipUp can be attributed to random noise.
With Adam, different step size settings introduced different sensitivities to the reward scale. The most stable setting for step size was 0.175. 
With step size 0.15, its performance dropped significantly when the rewards were multiplied by 1000. 
On the other hand, with step size 0.2, the performance dropped when the rewards were divided by 1000. 
Overall,
the performance of ClipUp was consistent across fitness scales while that of Adam was not.

\subsection{MuJoCo continuous control tasks}

Next we consider the continuous control tasks \texttt{Walker2d-v2} and \texttt{Humanoid-v2} defined in the Gym \citep{brockman2016} library, simulated using the MuJoCo \citep{todorov2012} physics engine.
The goal in these tasks is to make a robot walk forward.
In \texttt{Walker2d-v2}, the robot has a two-legged simplistic skeleton based on  \citet{erez2011}.
In \texttt{Humanoid-v2}, originally by \citet{tassa2012}, the robot has a much more complex humanoid skeleton.

Previous studies \citep{rajeswaran2017,mania2018} have demonstrated that a linear policy is sufficient to solve these tasks, and therefore we also adopt this approach.
The policy has the form $\VECVAR{act} = \VECVAR{obs} \cdot \MAT{W} + \VEC{b}$, where $\VEC{W}$ is a weight matrix, and $\VEC{b}$ is a bias vector.
In total, this results in 108 optimization variables for \texttt{Walker2d-v2}, and 6409 optimization variables for \texttt{Humanoid-v2}.

In these RL environments, by default the agents are rewarded a certain amount of ``alive bonus'' at each simulator timestep for not falling. 
\citet{mania2018} reported that this alive bonus causes the optimization to be driven towards agents that keep standing still to collect this alive bonus and not learning to walk. 
We experienced the same issue in our experiments, and therefore removed these alive bonuses.

Both tasks were solved using both PGPE+Adam and PGPE+ClipUp, each with 30 runs.
Each experiment run was configured to last for 500 iterations for \texttt{Walker2d-v2}, and 1000 iterations for \texttt{Humanoid-v2}.
The hyperparameters are reported in tables \ref{tbl:hyperwalker2d} and \ref{tbl:hyperhumanoid}.

\begin{table}
    \caption{Hyperparameters used for \texttt{Walker2d-v2}}
    \label{tbl:hyperwalker2d}
    \centering
    \begin{tabular}{c | c | c}
        \toprule
        PGPE & ClipUp & Adam \\
        \midrule
        $
        \begin{array}{rcl}
            \lambda &=& 100 \quad\quad
            \lambda^{\text{max}} \,=\, 800 \\
            T &=& \lambda \cdot 1000 \cdot (3/4) = 75000 \,\,^{(a)} \\
            \Omega &=& 0.1 \\
            r &=& v^{\text{max}} \cdot 15 \,=\, 0.225 \\
            \text{ObNorm} &=& \text{Yes} \quad\quad
            \text{Rank}^{(b)} \,=\, \text{Yes}
        \end{array}
        $
        &
        $
        \begin{array}{rcl}
            v^{\text{max}} &=& \num{1.5e-2} \\
            \alpha &=& v^{\text{max}} / 2 \,=\, \num{7.5e-3} \\
            m &=& 0.9 \\
        \end{array}
        $
        &
        $
        \begin{array}{rcl}
            \alpha &=& \num{4e-3} \,\,^{(c)} \\
            \beta_1 &=& 0.9 \\ 
            \beta_2 &=& 0.999 \\
            \epsilon &=& \num{1e-8} \\
        \end{array}
        $ \\
        \bottomrule
    \end{tabular}
    
    $ $
    
    \begin{tabular}{c p{14cm}}
    (a) & {\footnotesize 75\% of the population size times the maximum episode length, meaning on average 75\% of the solutions in a population must finish their assigned 1000-step episodes to the end, otherwise the population size is increased.} \\
    (b) & {\footnotesize 0-centered solution ranking} \\
    (c) & {\footnotesize Tuned in} 
        {\footnotesize $\{\num{4e-4}, \num{5e-4}, ..., \num{9e-4}, \num{1.2e-3}, \num{1.3e-3}, ..., \num{1.8e-3}, \num{2e-3}, \num{3e-3}, ..., \num{6e-3}\}$ (10 runs for each)} \\
    \end{tabular}
\end{table}
\begin{table}
    \caption{Hyperparameters used for \texttt{Humanoid-v2}}
    \label{tbl:hyperhumanoid}
    \centering
    \begin{tabular}{c | c | c}
        \toprule
        PGPE & ClipUp & Adam \\
        \midrule
        $
        \begin{array}{rcl}
            \lambda &=& 200 \quad\quad
            \lambda^{\text{max}} \,=\, 3200 \\
            T &=& \lambda \cdot 1000 \cdot (3/4)=150000 \\
            \Omega &=& 0.1 \\
            r &=& v^{\text{max}} \cdot 18 \,=\, 0.27 \\
            \text{ObNorm} &=& \text{Yes} \quad\quad
            \text{Rank} \,=\, \text{Yes} \\
        \end{array}
        $
        &
        $
        \begin{array}{rl}
            v^{\text{max}}=& \num{1.5e-2} \\
            \alpha=& v^{\text{max}}/2=\num{7.5e-3}\,\,^{(a)} \\
            m=& 0.9 \\
        \end{array}
        $
        &
        $
        \begin{array}{rcl}
            \alpha &=& \num{6e-4} \,\,^{(b)} \\
            \beta_1 &=& 0.9 \\ 
            \beta_2 &=& 0.999 \\
            \epsilon &=& \num{1e-8} \\
        \end{array}
        $ \\
        \bottomrule
    \end{tabular}
    
    $ $
    
    \begin{tabular}{c p{14cm}}
    (a) & {\footnotesize Can slightly boost the performance if increased to \num{1.1e-2} (see \Cref{sec:tunestepsize})} \\
    (b) & {\footnotesize Tuned in the set $\{\num{4e-4}, \num{5e-4}, ..., \num{9e-4}\}$ (10 runs each)}
    \end{tabular}
\end{table}
\begin{figure}
    \centering
    \includegraphics[trim={0cm 0 0.65cm 0}, clip, width=0.48\textwidth]{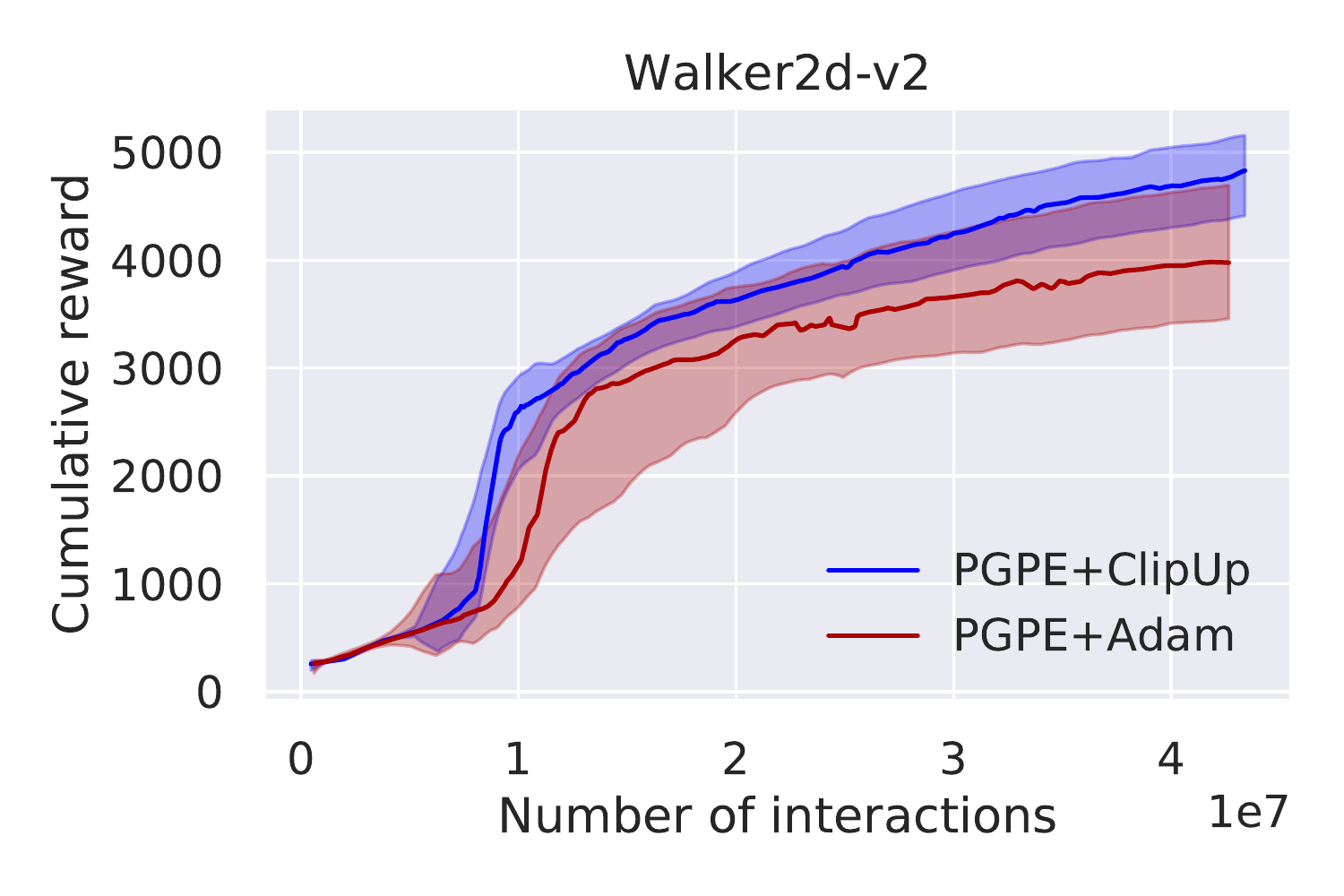}
    \includegraphics[trim={0cm 0 0.65cm 0}, clip, width=0.48\textwidth]{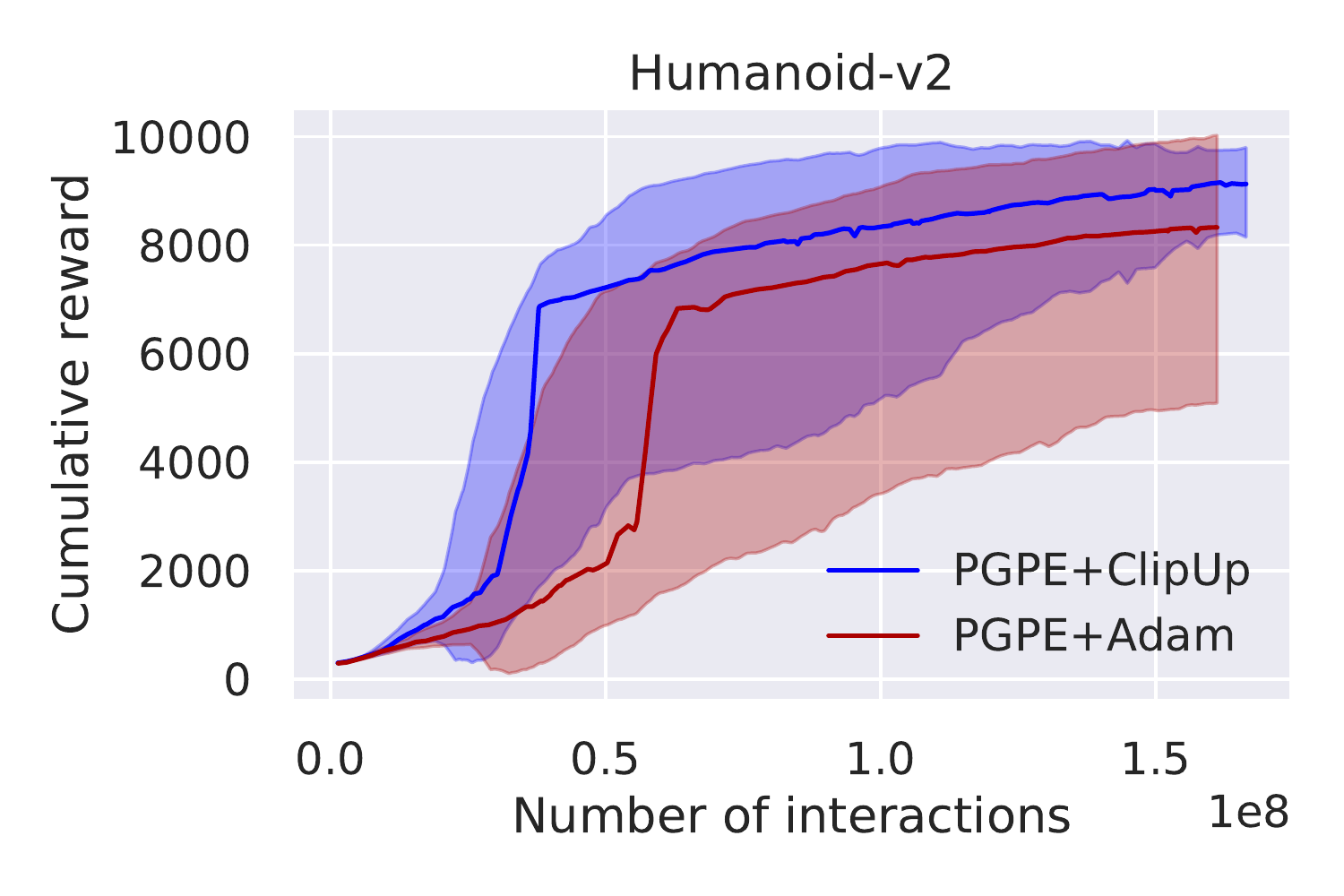}
    
    \caption{Comparison of training curves for PGPE+ClipUp and PGPE+Adam on Walker2d-v2 and Humanoid-v2 based on 30 runs each.
    Each run's reported cumulative reward at a time is the result of 16 re-evaluations averaged.
    The x-axis represents the number of interactions made with the simulator (i.e. the number of simulator timesteps). Dark lines mark the median cumulative reward values. The shaded regions are bounded by the mean $\pm$ standard deviations of the cumulative rewards.}
    \label{fig:mujoco}
\end{figure}

The learning curves obtained with ClipUp and Adam are compared in \Cref{fig:mujoco}.
In both cases, the eventual performance of the two algorithms was very similar, but ClipUp jumped to high cumulative rewards earlier for \texttt{Humanoid-v2}.
Both algorithms scored over 6000 on \texttt{Humanoid-v2}, clearing the official threshold for solving the task.

\subsection{PyBullet Humanoid}

\begin{table}[t]
    \caption{Hyperparameters used for \texttt{HumanoidBulletEnv-v0}}
    \label{tbl:hyperpybhmn}
    \centering
    \begin{tabular}{c | c}
        \toprule
        PGPE & ClipUp \\
        \midrule
        $
        \begin{array}{rcl}
            \lambda &=& 10000 \quad\quad
            \lambda^{\text{max}} \,=\, 80000 \\
            T &=& \lambda \cdot 200 \cdot (3/4) \,=\, 1500000 \\
            \Omega &=& 0.1 \\
            r &=& v^{\text{max}} \cdot 15 \,=\, 2.25 \\
            \text{ObNorm} &=& \text{Yes} \quad\quad
            \text{Rank} \,=\, \text{Yes} \\
        \end{array}
        $
        &
        $
        \begin{array}{rcl}
            v^{\text{max}} &=& 0.15 \\
            \alpha &=& v^{\text{max}} / 2 \,=\, 0.075 \\
            m &=& 0.9 \\
        \end{array}
        $
        \\
        \bottomrule
    \end{tabular}
\end{table}
\begin{figure}[t]
    \centering
    \includegraphics[width=0.625\textwidth]{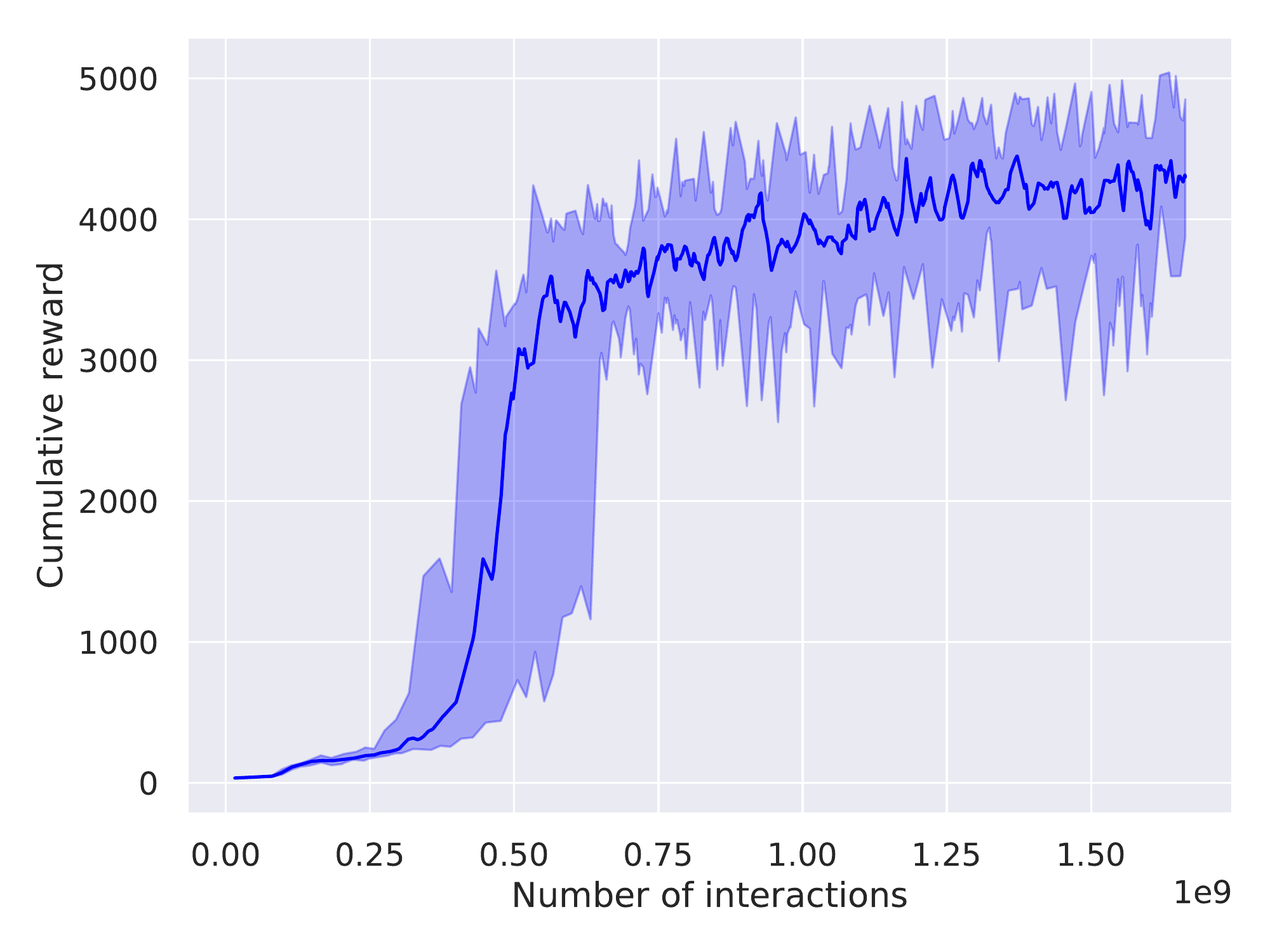}
    \caption{PGPE+ClipUp solves the challenging HumanoidBulletEnv-v0 control task.
    The dark line marks the median evaluated cumulative reward over 10 runs, and the shaded region is bounded by the minimum and the maximum cumulative rewards.}
    \label{fig:pybullethumanoid}
\end{figure}

As a stress test of ClipUp's utility, we attempted to use ClipUp and our proposed hyperparameter tuning scheme for reliably solving the challenging RL task labeled \texttt{HumanoidBulletEnv-v0}, defined in and simulated by the PyBullet \citep{coumans2019} library.
This task also involves teaching a model humanoid to walk forward, but as noted by the author of PyBullet \citep{coumans2018note}, this version of the task is much harder than its MuJoCo counterpart.
Perhaps because of this mentioned difficulty, successful results for it are rarely reported.

As in MuJoCo experiments, the default alive bonus for this task was removed.
In addition, trajectory length upper bound was decreased from 1000 timesteps to 200 timesteps, since the hardest part of the task is starting a forward gait (the terrain is flat and there are no randomized traps). 
We used a neural network policy with a single hidden layer of 64 neurons.
The architecture of this policy can be formulated as
$\VECVAR{act} = \text{tanh}(\VECVAR{obs} \cdot \MAT{W}_1 + \VEC{b}_1) \cdot \MAT{W}_2 + \VEC{b}_2$, where $\MAT{W}_i$ and $\VEC{b}_i$ represent the $i$-th layer's weight matrix and bias vector, respectively.
This policy results in 3985 optimization variables.
We configured PGPE to run for 1000 iterations.
The hyperparameters are reported in \Cref{tbl:hyperpybhmn}.

Each run was on an Amazon EC2 m4.16xlarge instance (64 vCPUs).
The performance of PGPE+ClipUp vs.\@ number of environment interactions is shown in \Cref{fig:pybullethumanoid}.
The median cumulative reward reached 3500 (the solving threshold defined by \citet{klimov2017}) within $0.75\times10^9$ steps (about 15 hours of training), and then stayed mostly above 4000 after $10^9$ steps (about 24 hours).
This result confirms that despite its simplicity, ClipUp is effective at solving difficult control problems.

\subsection{Effect of tuning the step size}
\label{sec:tunestepsize}

The experiments described in this section so far followed the hyperparameter tuning heuristics mentioned in \Cref{sec:tuning}.
These heuristics serve to save a practitioner time and effort in obtaining a useful set of working hyperparameters.
However, when additional resources are available, we have observed that the step size can be tuned to higher values to obtain slight performance boost and/or faster training.
As an example, we retrained the \texttt{Humanoid-v2} agent, while using the same the radius and $v^{\text{max}}$ values from \Cref{sec:experiments}, but with the step size $\alpha$ increased from $\num{7.5e-3}$ to $\num{1.1e-2}$ (larger values led to reduced performance).
It was found that although the difference in the mean (over 30 runs) of the cumulative rewards after 1000 generations was not significant, higher $\alpha$ values solved the task (i.e. reached above the cumulative reward threshold of 6000) quicker.
On average, moving $\alpha$ from $\num{7.5e-3}$ to $\num{1.1e-2}$ decreased the number of required simulator interactions by 19.11\% ($\pm$ 16.84\%, with a confidence of 90\%).

\section{Discussion}
\label{sec:behavior}

ClipUp adds two ingredients to stochastic gradient descent with momentum: gradient normalization and update clipping.
The importance of normalized gradients is easier to understand based on sensitivity to fitness scale (\Cref{subsec:scale}) and bringing the learning rate to a more natural scale, but the clipping operator appears relatively ad-hoc at first glance.
In this section we present experimental observations that demonstrate how and why clipped updates are beneficial.
Before proceeding, note that since $v^{\text{max}}$ is a hyperparameter, ClipUp's theoretical properties would be that of normalized gradient descent with momentum.\footnote{Such a ``NoClip'' optimizer was used by \citet{salomon1998}.}
Here we are more interested in the behavior of optimizers in practice, and in particular, that of the magnitude of the updates (the \emph{speed}) during the course of optimization.

\begin{figure}[t]
    \centering
    \begin{tabular}{c c}
        \includegraphics[width=0.46\textwidth]{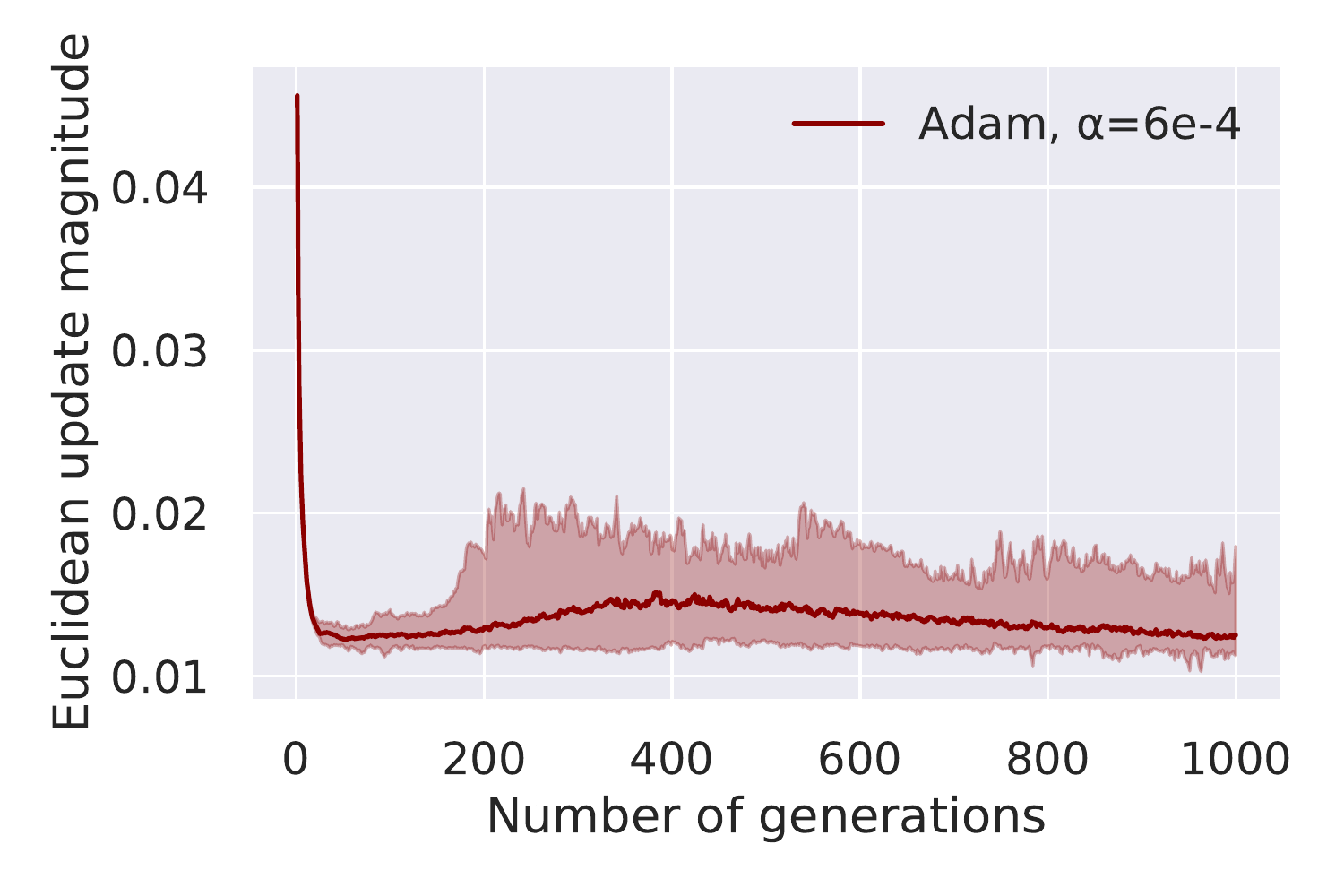}
         &
        \includegraphics[width=0.46\textwidth]{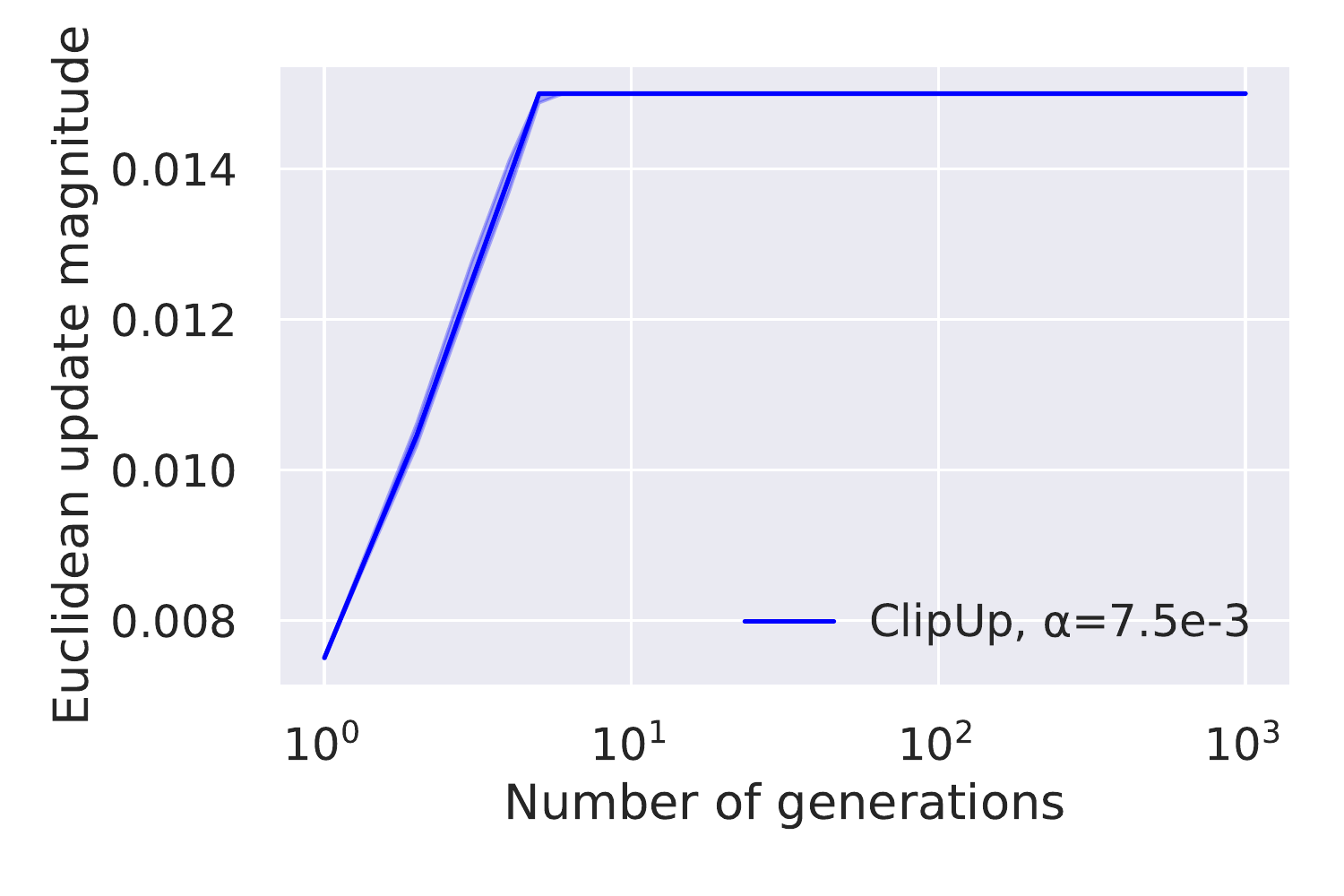}
    \end{tabular}
    \caption{Update magnitudes at each generation while solving Humanoid-v2 using Adam and ClipUp optimizers, for experiments summarized in \Cref{fig:mujoco}.
    The dark curves represent the medians, and shaded areas are bounded by the minimum and the maximum values.
    The key observation is that the update magnitudes typically settle to certain stable values, dependent on the hyperparameters.
    }
    \label{fig:humanoidUpdateMagnitudes}
\end{figure}

From our experiments with both Adam and ClipUp, two observations about the speed are notable: a) after an initial phase of adjustment, it typically ``settles'' around a certain value during optimization and b) this settled speed is \emph{indirectly} controlled by the optimizer hyperparameters.
The evolution of speed when training on \texttt{Humanoid-v2} using tuned hyperparameters for both optimizers is shown in \Cref{fig:humanoidUpdateMagnitudes}.
In the case of Adam, relatively large updates are done in the first few generations, and after a quick decay, the speed drops to values between 0.01 and 0.02.
With ClipUp, the first update's magnitude is $\num{7.5e-3}$ (the step size, $\alpha$, which can also be interpreted as the \emph{initial speed}).
The speed then grows until it hits the maximum speed $v^{\text{max}}=\num{1.5e-2}$.
Despite differences in initial behavior, both optimizers settle around similar speeds.
We believe that it is because of settling around the ``correct'' speed (as a result of the selected hyperparameters) that both optimizers are able to train successful \texttt{Humanoid-v2} policies. 

The main idea behind ClipUp is to \emph{directly} control this final speed of training via $v^{\text{max}}$. 
To be useful in practice, $v^{\text{max}}$ should not be much larger than $\alpha$, otherwise the clipping threshold will never be reached.
For example, for the tasks considered in this study, we found that the initial tuning rule $\alpha=v^{\text{max}}/2$ in \Cref{sec:tuning} typically set the two hyperparameters to be close enough, and the optimizer indeed settled its update magnitudes around $v^{\text{max}}$.
As long as this condition is satisfied, $v^{\text{max}}$ is the main hyperparameter controlling the behavior of ClipUp.
Note that even after the speed reaches $v^{\text{max}}$ and the remaining updates have the same magnitude (as is the case in \Cref{fig:humanoidUpdateMagnitudes}), the moving averaging mechanism of momentum still remains in effect for the \emph{directions} of the updates.

\subsection{Utility of the clipping operator}
\label{sec:clippingeffect}

Depending on the values of other hyperparameters, the clipping operator is not always essential for successful training, i.e. one can disable clipping while keeping other hyperparameters the same without a drop in performance.
However, clipping becomes increasingly important in certain cases that are very relevant in practice.
Here we identify two such cases where the goal is to reduce the time and resources required to obtain a trained solution:
\inum{i} increased step sizes, as done in \Cref{sec:tunestepsize}; and
\inum{ii} lower population sizes.
We directly compare ClipUp to \emph{NoClip}, which is exactly the same optimizer but with clipping disabled (or set to a very large value).

\begin{figure*}[tbp]
    \centering
    \begin{subfigure}[b]{0.495\textwidth}
        \includegraphics[width=\textwidth]{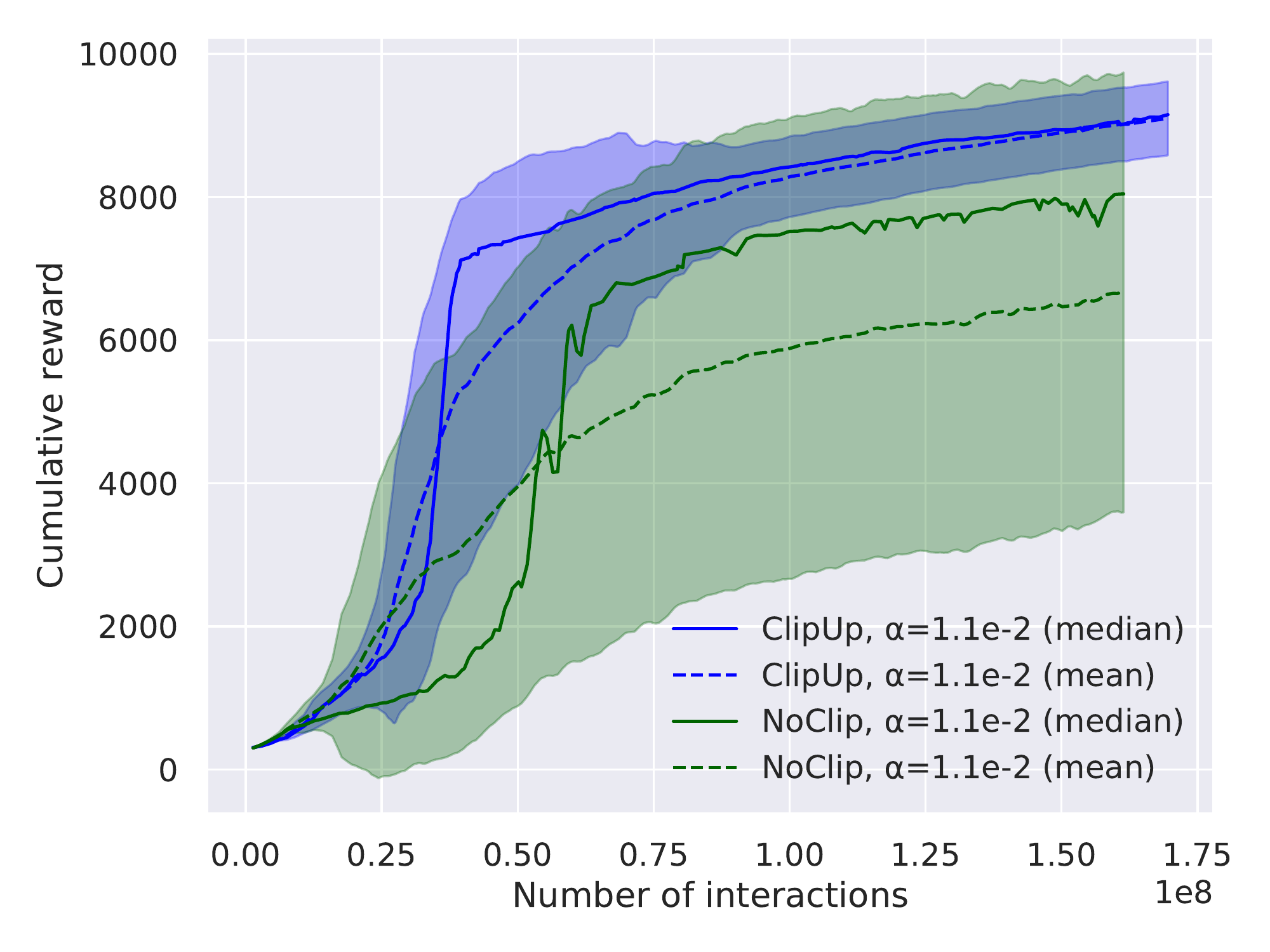}
        \caption{}
        \label{fig:humanoidMoreLrClipVsNoClip}
    \end{subfigure}
        \begin{subfigure}[b]{0.495\textwidth}
        \includegraphics[width=\textwidth]{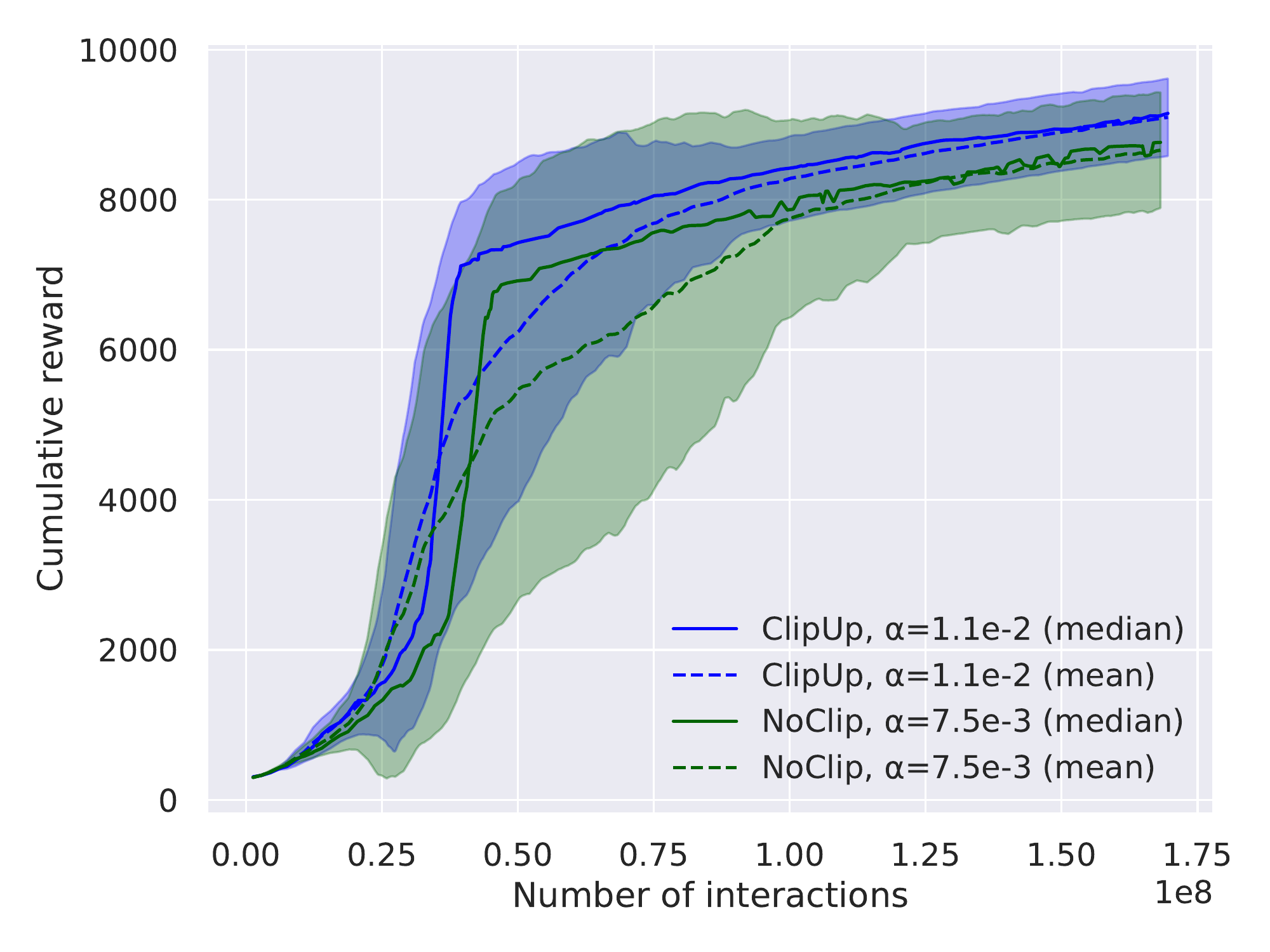}
        \caption{}
        \label{fig:humanoidMoreLrClipVsSlowNoClip}
    \end{subfigure}
    
    \caption{
        Use of the clipping operator leads to improved/faster results.
        These plots compare 
        \textbf{a)} PGPE+ClipUp vs. PGPE+NoClip ($\alpha=\num{1.1e-2}$), and
        \textbf{b)} PGPE+ClipUp ($\alpha=\num{1.1e-2}$) vs. PGPE+NoClip ($\alpha=\num{7.5e-3}$) on Humanoid-v2.
        Shaded areas are bounded by the means $\pm$ the standard deviations over 30 runs.
    }
    \label{fig:humanoidMoreLr}
\end{figure*}

\tinysection{Increased step sizes}
As mentioned in \Cref{sec:tunestepsize}, increasing the ClipUp step size from its default setting can reduce the training time significantly for some problems.
Does clipping play a role in enabling the use of higher step sizes?

\begin{figure*}[tbp]
    \centering
    \begin{subfigure}[b]{0.66\textwidth}
        \includegraphics[width=0.495\textwidth]{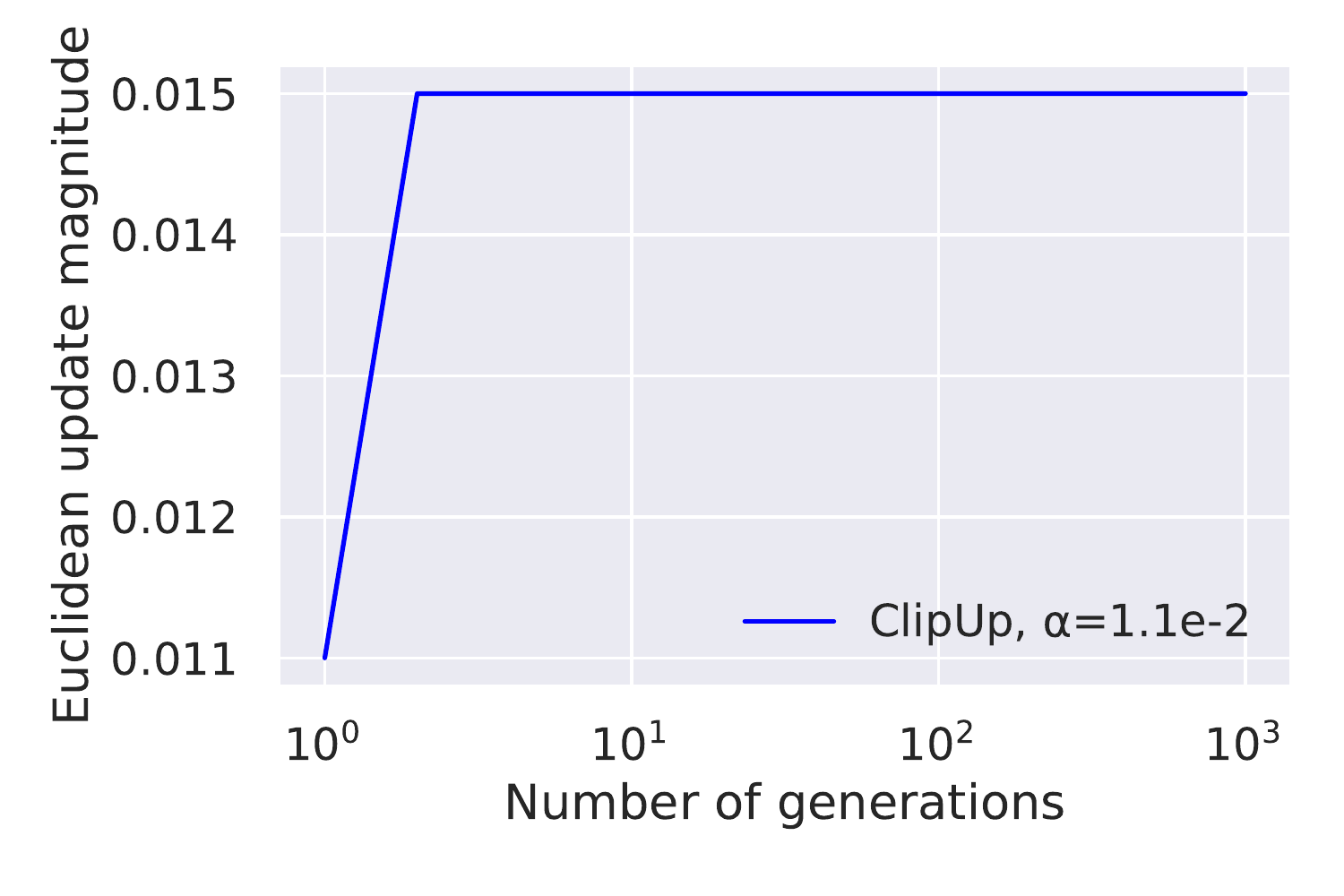}
        \includegraphics[width=0.495\textwidth]{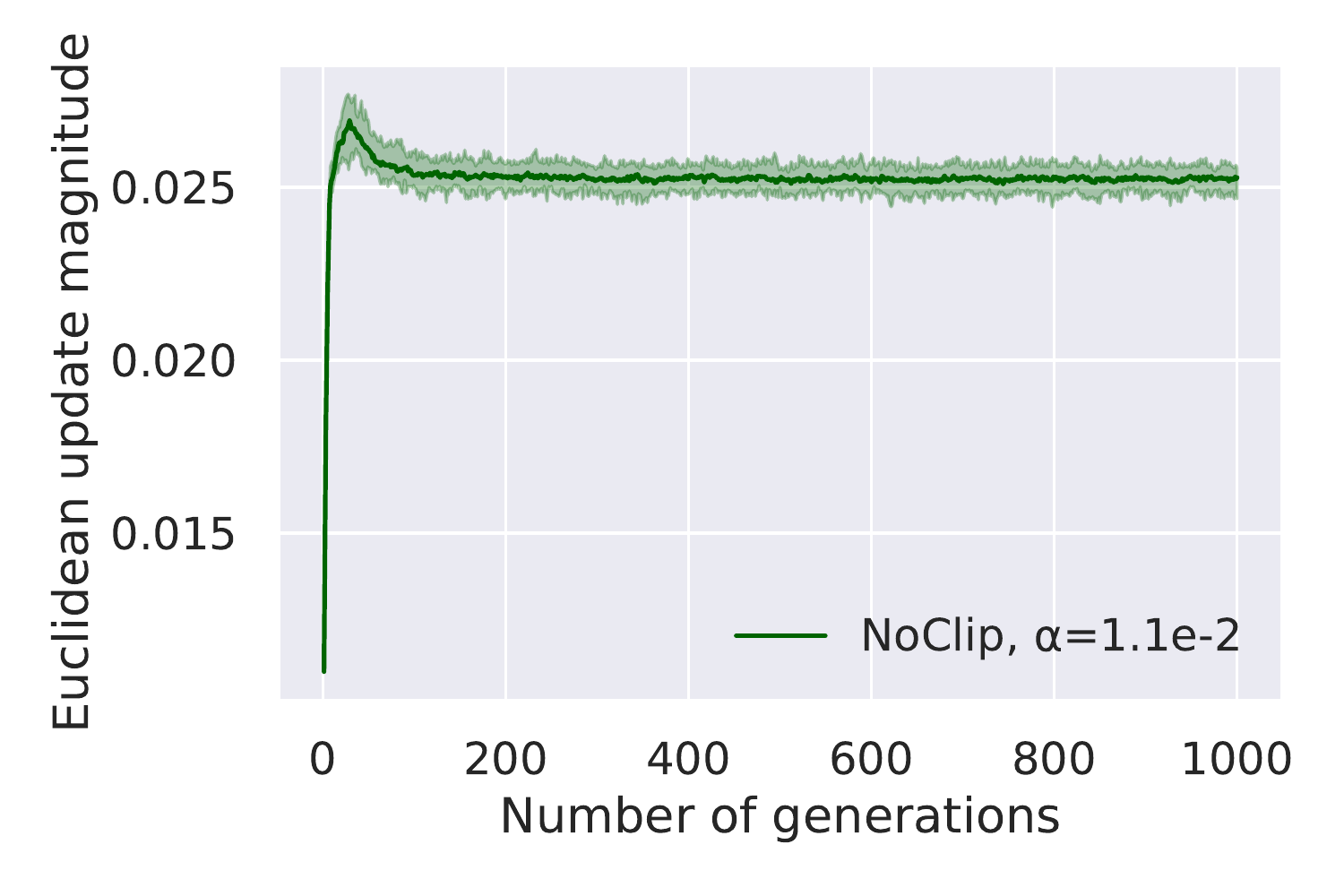}
        \caption{}
        \label{fig:humanoidMoreLrClipUpVsNoClipMagnitudes}
    \end{subfigure}
        \begin{subfigure}[b]{0.33\textwidth}
        \includegraphics[width=\textwidth]{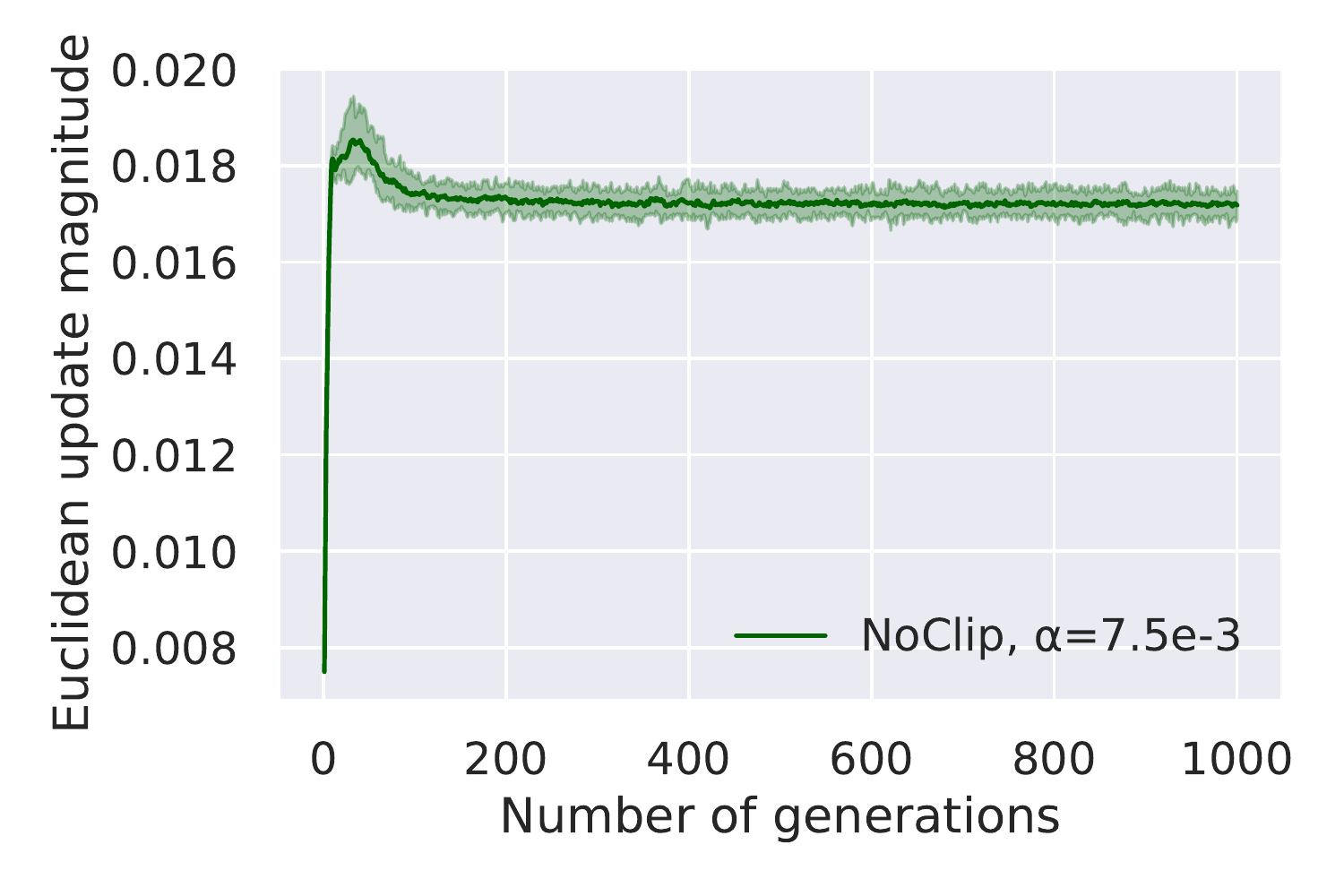}
        \caption{}
        \label{fig:humanoidLessLrNoClipMagnitudes}
    \end{subfigure}
    
    \caption{
        Update magnitudes for 
        \textbf{a)} ClipUp and NoClip ($\alpha=\num{1.1e-2}$),
        \textbf{b)} NoClip $\alpha=\num{7.5e-3}$ on Humanoid-v2.
        The shaded areas are bounded by the minimum and the maximum values, and the dark curves represent the medians over 30 runs.
        We see that update clipping directly controls the update magnitudes for ClipUp, while for NoClip, updates settle at larger magnitudes and lead to reduced performance.
    }
    \label{fig:humanoidMoreLrMagnitudes}
\end{figure*}

\Cref{fig:humanoidMoreLrClipVsNoClip} shows a comparison between the performance of ClipUp and NoClip using 30 runs each with the higher $\alpha=\num{1.1e-2}$ identified previously.
The median performance of PGPE+NoClip is clearly worse than PGPE+ClipUp, and the gap is even larger in terms of mean performance, due to some runs performing much worse throughout training.
Indeed, further analysis of these results showed that 8 of the PGPE+NoClip agents were below the solving threshold of the task while all 30 PGPE+ClipUp agents cleared the solving threshold.
\Cref{fig:humanoidMoreLrClipUpVsNoClipMagnitudes} shows that the update magnitudes of NoClip reached above $\num{2.5e-2}$, which is significantly larger than $\num{1.5e-2}$, the maximum speed used by ClipUp.

One may speculate that while NoClip is outperformed by ClipUp at higher step sizes due to lack of clipping, it might yield better performance with lower learning rates that grow the update size more stably.
We test this hypothesis by comparing ClipUp with $\alpha=\num{1.1e-2}$ to NoClip with the original $\alpha=\num{7.5e-3}$.
As shown by results in \Cref{fig:humanoidMoreLrClipVsSlowNoClip}, a lower step size indeed works better and eliminates the poorly performing runs.
However, ClipUp still maintains a slight advantage overall.
On average, compared to NoClip, ClipUp with the higher $\alpha$ decreases the number of required simulator interactions to reach the solving threshold by 17.05\% ($\pm$ 15.34\%, with a confidence of 90\%).
\Cref{fig:humanoidLessLrNoClipMagnitudes} shows the update magnitudes for NoClip with the lower learning rate, and we find that they settle to values between $\num{1.6e-2}$ and $\num{1.8e-2}$, i.e. much closer to ClipUp's $v^{\text{max}}=\num{1.5e-2}$.

\begin{figure*}[tbp]
    \centering
    \begin{subfigure}[b]{0.635\textwidth}
        \includegraphics[width=\textwidth]{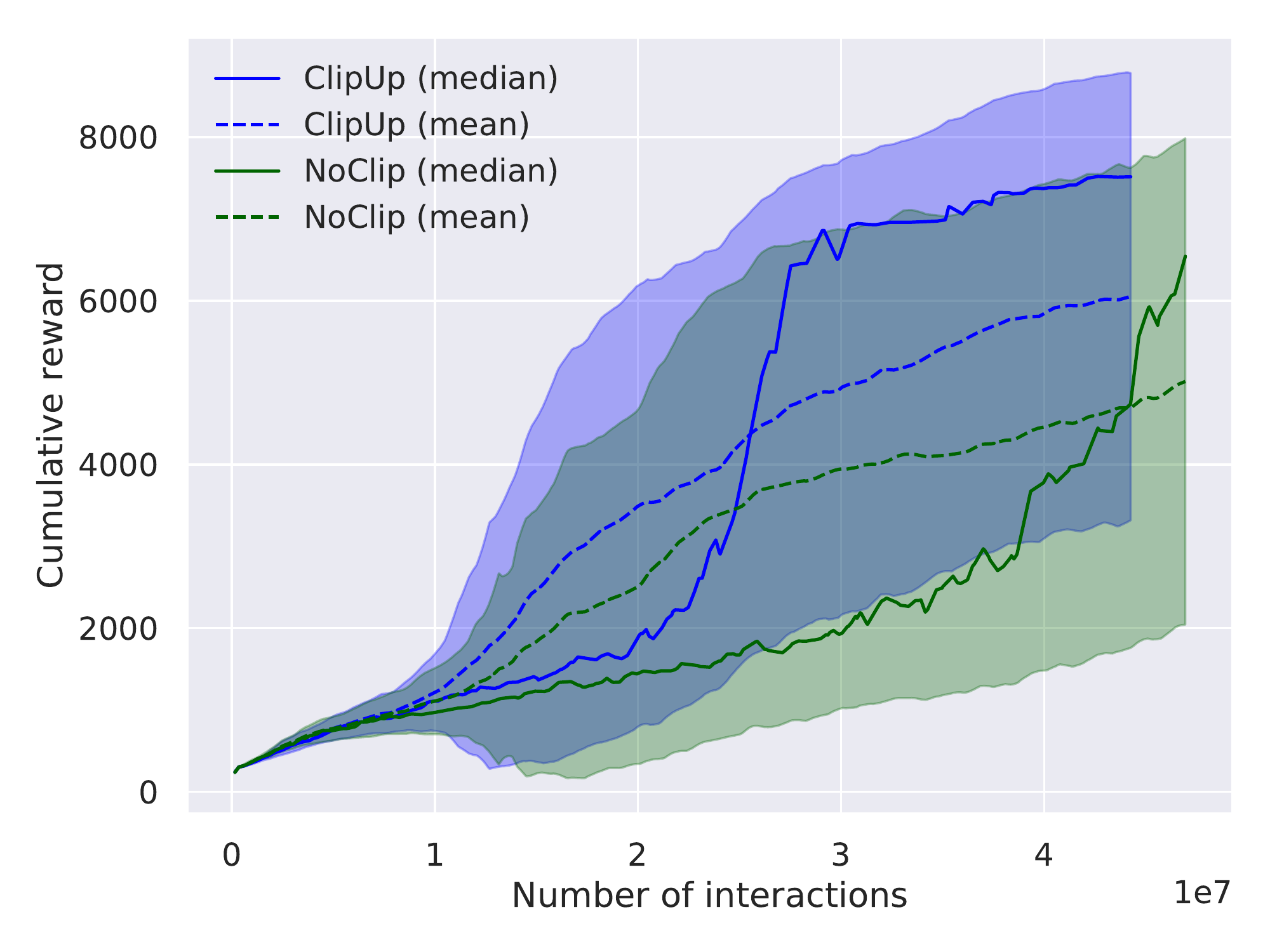}
        \caption{}
        \label{fig:humanoidLittlePopClipVsNoclip}
    \end{subfigure}
        \begin{subfigure}[b]{0.355\textwidth}
        \includegraphics[width=\textwidth]{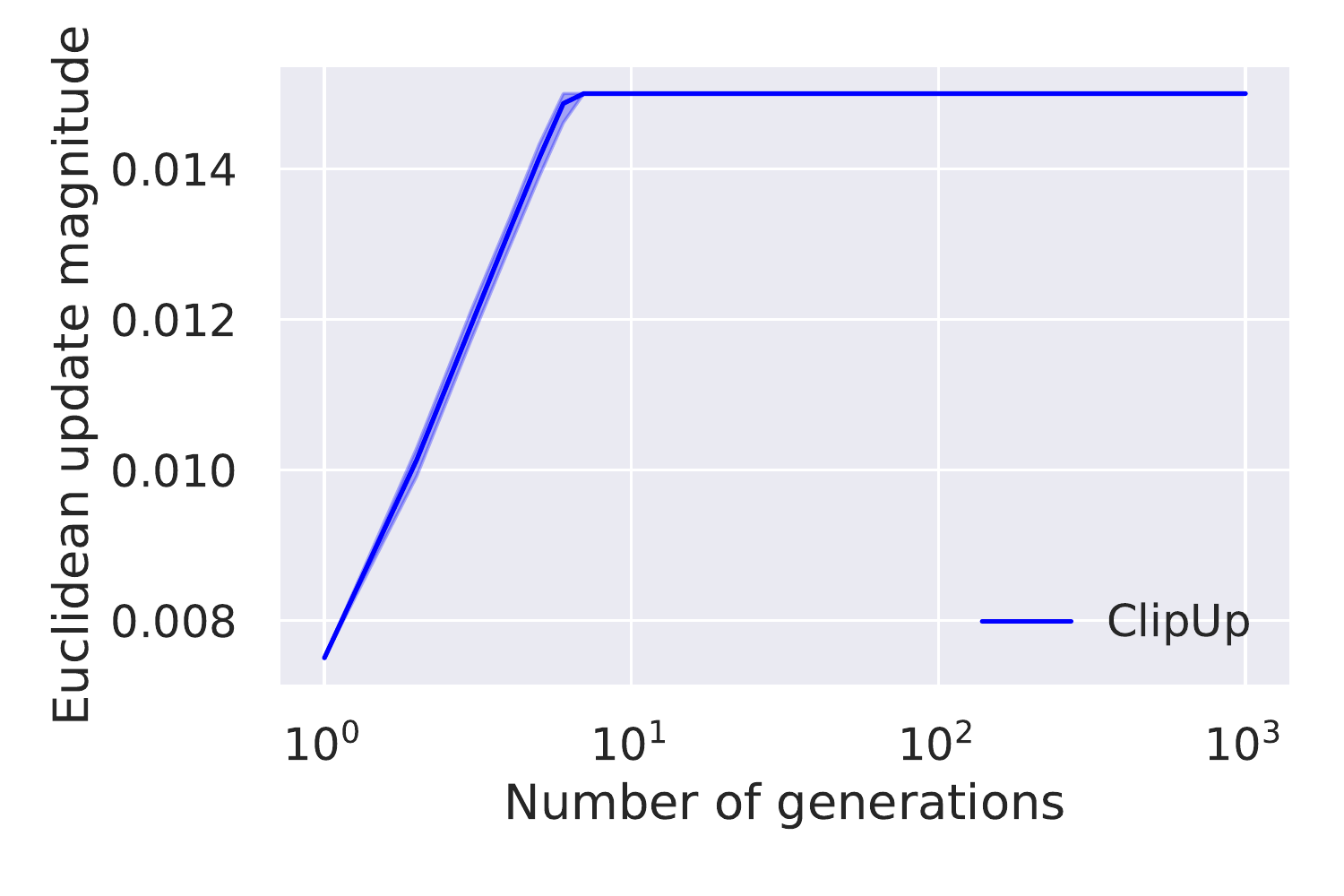}
        \includegraphics[width=\textwidth]{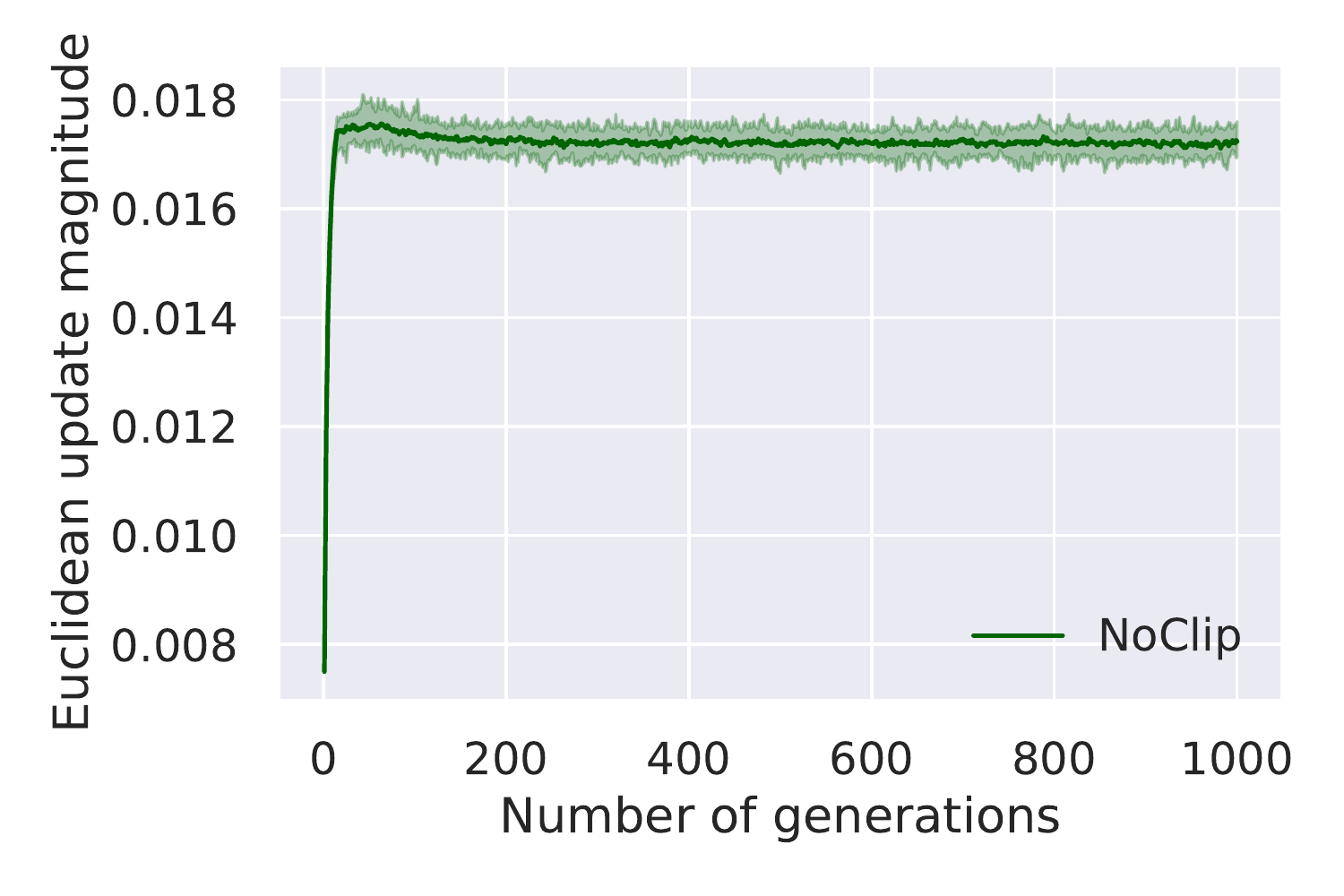}
        \caption{}
        \label{fig:humanoidLittlePopMagnitudes}
    \end{subfigure}
    
    \caption{
        Comparison of ClipUp and NoClip on Humanoid-v2 using 30 runs, with decreased population sizes.
        These plots compare
        \textbf{a)}
        cumulative rewards (shaded regions bounded by the means $\pm$ the standard deviations), and 
        \textbf{b)}
        update magnitudes (shaded regions bounded by the minimum and maximum values).
        Update clipping becomes critical when the gradient estimates are unreliable due to low population sizes.
    }
    \label{fig:humanoidLittlePop}
\end{figure*}

\tinysection{Decreased population sizes}
When computational resources are limited, a practitioner may choose to solve the task at hand with decreased population sizes.
However, at lower population sizes, the gradient estimates become increasingly noisy.
Does clipping play a role in mitigating the effect of this noise?

We compared the behaviors of ClipUp and NoClip on \texttt{Humanoid-v2}, with $\lambda$ decreased from 200 to 80, $\lambda^{\text{max}}$ decreased from 3200 to 400, and $T$ decreased from 150000 to 60000.
Other hyperparameters from \Cref{sec:experiments} were reused.
In \Cref{fig:humanoidLittlePopClipVsNoclip}, it can be seen that both ClipUp and NoClip suffered from noisy gradient estimations resulting in a large increase in performance variance.
Nevertheless, ClipUp had an advantage based on both the mean and median performance.
\Cref{fig:humanoidLittlePopMagnitudes} plots the update magnitudes during training and shows that NoClip settled around higher update magnitudes compared to ClipUp, similar to the setting with higher population size (compare \Cref{fig:humanoidLessLrNoClipMagnitudes}).
This indicates that careful clipping of the update becomes more critical when the gradient is less reliable.

\section{Conclusions}
\label{sec:conclusions}

ClipUp builds upon \emph{normalized gradient descent}, which was recently shown to converge faster than fixed-step gradient descent under mild conditions \citep{zhang2020}.
While a general theoretical analysis of normalized gradient descent with momentum is an open problem for future research, our aim in this paper was to develop a simple and effective optimizer to aid in the practice of distribution-based evolutionary RL.

We argued that using ClipUp in this context is intuitive, mainly thanks to the following:
\begin{itemize}
    \item its step size and the maximum speed (the two main hyperparameters) are relatable to the magnitude of the mutation one would like to apply on the current solution;
    \item its step size is robust to the fitness scale of the problem;
    \item one can tune a single hyperparameter in practice --- the maximum speed --- and determine the step size and the initial search distribution's radius by following simple heuristic rules (e.g. step size as half the maximum speed, and radius about 10 to 20 times the maximum speed).
\end{itemize}
These properties improve a practitioner's experience when applying distribution-based search to RL problems.
In addition, we found ClipUp to be competitive against the well-known Adam optimizer on the MuJoCo continuous control tasks \texttt{Walker2d-v2} and \texttt{Humanoid-v2}.
Finally, we showed that PGPE with ClipUp can successfully solve the \texttt{HumanoidBulletEnv-v0} benchmark, demonstrating its applicability to highly challenging control tasks.

Although we used PGPE in our experiments, ClipUp can be used with any evolution strategy variant where the solution update is in the form of gradient estimation (e.g. the algorithm used by \citet{salimans2017}).
While we hope ClipUp will be a valuable tool for practitioners, our broader goal is to encourage more investigations into optimization \emph{strategies} that take a practitioner's perspective into account, in addition to optimization algorithms that lie at their heart and enjoy good theoretical properties.

\bibliographystyle{humannat}
\bibliography{optimizers}

\clearpage
\appendix

\section{Software tools}
We used the following software to implement the mentioned algorithms and running our experiments:
\begin{itemize}
\setlength\itemsep{-1pt}
    \item Python (the reference ``CPython'' implementation, v3.6) \citep{vanrossum2009}
    \item NumPy (v1.18.1) \citep{oliphant2006}
    \item SciPy (v1.1.0) \citep{virtanen2020}
    \item gym (v0.17.0) \citep{brockman2016}
    \item PyTorch (v1.4.0) \citep{paszke2019}
    \item PyBullet (v2.6.6) \citep{coumans2019}
    \item ray (v0.7.2) \citep{moritz2018}
    \item sacred (v0.8) \citep{greff2017}
\end{itemize}

\section{Example Python Implementation of ClipUp}

\lstset{
    language=Python,
    numbers=left,
    numberstyle=\tiny\color{gray},
    basicstyle=\small\ttfamily,
    numbersep=4pt,
    keywordstyle=\color{Fuchsia},
    stringstyle=\color{Brown}
}

\begin{lstlisting}
    import numpy as np
    
    class ClipUp:
        @staticmethod
        def clip(x: np.ndarray, max_magnitude: float) -> np.ndarray:
            magnitude = np.sqrt(np.sum(x * x))
            if magnitude > max_magnitude:
                ratio = max_magnitude / magnitude
                return x * ratio
            else:
                return x
    
        def __init__(self,
                     solution_length: int,
                     stepsize: float,
                     momentum: float,
                     max_speed: float):
    
            self.velocity = np.zeros(solution_length, dtype="float32")
            self.stepsize = stepsize
            self.momentum = momentum
            self.max_speed = max_speed
    
        def compute_ascent(self, gradient: np.ndarray) -> np.ndarray:
            grad_magnitude = np.sqrt(np.sum(gradient * gradient))
            gradient = gradient / grad_magnitude
    
            step = gradient * self.stepsize
    
            self.velocity = self.momentum * self.velocity + step
            self.velocity = self.clip(self.velocity, self.max_speed)
    
            return self.velocity
\end{lstlisting}

\section{PGPE with Reward normalization}
Following \cite{ha2017,ha2019}, we use zero-centered fitness ranking in PGPE, in place of the reward normalization in the original PGPE definition \citep{sehnke2010}.
For completeness, we note that the original reward normalization would result in the following gradient estimates (compare \Cref{alg:pgpe}):

\begin{equation*}
    \nabla_{x_k} \gets \frac{1}{|D_k|} \cdot \sum_{(d^+,d^-) \in D_k} \bigg[ \frac{f(d^+) - f(d^-)}{2} \cdot \frac{1}{f^{\text{max}} - \frac{f(d^+) + f(d^-)}{2}} \cdot (d^+-x_k) \bigg]
\end{equation*}

\begin{equation*}
    \nabla_{\sigma_k} \gets
    \frac{1}{|D_k|}
    \cdot
    \sum_{(d^+,\,d^-) \in D_k}
    \Bigg[
    \bigg(
    \frac{1}{f^{\text{max}}-\widetilde{f}}
    \bigg)
    \cdot
    \bigg(
    \frac{f(d^+) + f(d^-)}{2} - \widetilde{f}
    \bigg)
    \cdot
    \bigg(
    \frac{(d^+-x_k)^2 - (\sigma_k)^2}{\sigma_k}
    \bigg)
    \Bigg]
\end{equation*}
where $f^\text{max}$ is the maximum cumulative reward an agent can get if such information is available, otherwise, the maximum cumulative reward achieved so far.

\end{document}